\pdfoutput=1

\documentclass[11pt]{article}

\usepackage{acl}

\usepackage{times}
\usepackage{latexsym}
\usepackage{tcolorbox}
\usepackage{multirow}
\usepackage{array}
\usepackage{wrapfig} 

\usepackage[T1]{fontenc}

\usepackage[utf8]{inputenc}

\usepackage{microtype}

\usepackage{inconsolata}

\usepackage{graphicx}

\usepackage{booktabs}       
\usepackage{amsfonts}       
\usepackage{amsmath}
\usepackage{nicefrac}       
\usepackage{microtype}      
\usepackage{xcolor}         
\usepackage{caption} 
\usepackage{subcaption} 
\usepackage{comment}

\usepackage{tabularx}
\usepackage{float} 

\title{Reinforcement Learning vs. Distillation: \\Understanding Accuracy and Capability in LLM Reasoning}

  \author{
  Minwu Kim\thanks{Equal contribution.} \thanks{Correspondence to \texttt{mwk300@nyu.edu}.} \quad Anubhav Shrestha\textsuperscript{*} \quad Safal Shrestha \quad Aadim Nepal \quad Keith Ross \\
  New York University Abu Dhabi
}

\begin{document}
\maketitle
\renewcommand{\thefootnote}{\fnsymbol{footnote}}

\renewcommand{\thefootnote}{\arabic{footnote}}
\begin{abstract}
Recent studies have shown that reinforcement learning with verifiable rewards (RLVR) enhances overall accuracy (pass@1) but often fails to improve capability (pass@$k$) of LLMs in reasoning tasks, while distillation can improve both. 
In this paper, we investigate the mechanisms behind these phenomena. 
First, we demonstrate that RLVR struggles to improve capability as it focuses on improving the accuracy of the easier questions to the detriment of the accuracy of the most difficult questions.
Second, we show that RLVR does not merely increase the success probability for the easier questions, but in our small model settings, produces quality responses that were absent in its original output distribution. In addition, we show these responses are neither noticeably longer nor feature more reflection-related keywords, underscoring the need for more reliable indicators of response quality.
Third, from the experiment distilling teacher responses to in-distribution problems, we find that capability does not always improve with distillation. We conjecture that capability improves only when new knowledge is introduced, whereas distilling reasoning patterns only improves accuracy but not capability, sacrificing performance on the most difficult questions, similar to RLVR.
Together, these findings offer a clearer understanding of how RLVR and distillation shape reasoning behavior in LLMs.\footnote{Code available at \url{https://github.com/minwukim/RLvsDistillation}}


\end{abstract}

\section{Introduction}

Large language models (LLMs) have made rapid progress in complex domains such as mathematics and programming. A key development is the emergence of reasoning models \cite{openai2024o1, deepseek2025r1, kimi2025k1.5}, which outperform conventional LLMs by employing more advanced reasoning strategies. Instead of relying on linear chains of thought (CoTs) \cite{wei2022chain}, these models exhibit non-linear behaviors such as subgoal formulation, verification, backtracking, and backward chaining \cite{gandhi2025cognitive, xiang2025towards}.


%

A central technique behind recent reasoning models is \emph{reinforcement learning with verifiable rewards (RLVR)} \cite{lambert2024tulu3, deepseek2025r1}. RLVR fine-tunes pre-trained LLMs using rewards based on whether the model’s output matches a ground-truth solution. Without explicit  
supervision, the model learns complex reasoning behaviors during training, making RLVR an effective fine-tuning method for reasoning tasks. In addition, {\em distillation} with responses from stronger reasoning models also provides strong performance, demonstrating that reasoning ability can be effectively transferred through supervised learning \cite{min2024imitate, deepseek2025r1, muennighoff2025s1, ye2025limo}.

It is well established that RLVR improves \textit{accuracy}—the probability of generating a correct answer, but whether it also improves \textit{capability}—the probability that a correct answer exists in the model’s output distribution—remains debated. Some studies suggest that, with sufficient compute and carefully matched training and test sets in skills and difficulty, RLVR can solve tasks that were previously unsolvable in certain domains \cite{liu2025prorl, setlur2025e3, sun2025omega}. Others, however, report that in more typical settings—where training and test sets contain heterogeneous problems with uncontrolled knowledge and difficulty—RLVR primarily amplifies existing reasoning rather than expanding capability \cite{dang2025assessing, wu2025invisibleleash, yue2025does, zhao2025echo}. By contrast, it has been observed that distillation improves both accuracy and capability \cite{yue2025does}. In this paper, we take a closer look at how RLVR and distillation shape mathematical reasoning in LLMs under typical settings, where training and test sets involve diverse problems with varying knowledge and difficulty.

Carrying out experiments with two models, Qwen2.5-1.5B-Math \cite{yang2024qwen25math} and Qwen2.5-3B \cite{hui2024qwen25}, we first demonstrate that RLVR does not improve capability because RLVR focuses on improving the accuracy of the less-difficult questions to the detriment of the accuracy of the most difficult questions, explaining why capability does not improve and can even decrease. We argue that this "sacrificing-difficult-problems" phenomenon is a direct consequence of the underlying policy-gradient algorithm in GRPO (Shao et al., 2024). We further find that RLVR does not merely increase the success probability for the easier questions, but in our small model settings produces responses that are more direct with fewer keywords. In addition, we find these responses are neither noticeably longer nor richer in reflection-related keywords (e.g., "wait", "alternatively"), underscoring the need for more reliable indicators of reasoning quality.

We next examine teacher distillation. A teacher model’s responses convey two main elements: (1) reasoning patterns and (2) domain knowledge. To disentangle their effects, we compare three models: the base model, the publicly released DeepSeek reasoning model, which is distilled on 800k responses from DeepSeek-R1 and likely incorporates substantial new knowledge, and our own reasoning-only model, trained only on teacher responses for questions where the base model is already able to produce correct answers. We find that both distilled models yield substantial accuracy gains, but only the DeepSeek model shows clear capability improvement. These results indicate that teacher distillation does not always expand capability, even when accuracy meaningfully improves. While further investigation is needed to confirm, we conjecture that this difference stems from whether new knowledge is introduced during distillation: introducing new knowledge may expand capability, whereas distilling only reasoning patterns improves accuracy but not capability. Interestingly, for the reasoning-only model, we also find that accuracy of the easier questions improves to the detriment of the most difficult questions, mirroring the RLVR.


Taken together, our findings provide a clearer picture of different dynamics in the model behavior during RLVR training and distillation, and offer insights into strategies for enhancing the fundamental abilities of LLMs.

\section{Related Work}\label{related_work}

\textbf{Training reasoning models.} RLVR has emerged as a key method for training LLMs to tackle complex reasoning tasks by generating long CoTs \cite{deepseek2025r1, lambert2024tulu3, openai2024o1}. It has shown strong performance across model sizes \cite{gandhi2025cognitive, hu2025openreasonerzero, liu2025understanding, xu2025phi4mini, yeo2025demystifying, zeng2025simplerl} and domains \cite{pan2025tinyzero, shrestha2025warmup, xie2025logicrl, zhang2025medrlvr}. Numerous RLVR variants have also been proposed to improve performance, data efficiency, and computational cost \cite{fatemi2025concise, liu2025understanding, shao2025spurious, shao2024deepseekmath, wang2025one, wang2025oneexample, yu2025dapo, zuo2025ttrl}. Distilling high-quality CoT data is another effective approach for enhancing LLM reasoning. Such data are obtained either by prompting large models \cite{distilling2024system2, zelikman2022star} or by human annotation of complex reasoning traces \cite{qin2024o1, xiang2025towards, ye2025limo}. A widely used strategy now involves distilling long CoT responses from RLVR-trained models into student models, often yielding substantial performance gains \cite{huang2024o1, min2024imitate, muennighoff2025s1, shrestha2025warmup}. Our work examines both RLVR and distillation, and evaluates how these two approaches differentially shape reasoning behavior in LLMs.

\noindent \textbf{Capability expansion in RLVR.} There is ongoing debate about whether RLVR develops genuinely new capabilities not already present in a model. Several works \cite{dang2025assessing, yue2025does, zhao2025echo} argue that RLVR merely amplifies correct reasoning already latent in the model. By contrast, ProRL demonstrates empirically that, given sufficient compute and diverse data, RLVR can enable models to solve previously unsolvable tasks in some domains-such as logic puzzles—suggesting the possibility of true capability expansion \cite{liu2025prorl}. OMEGA provides a more controlled analysis by carefully adjusting the knowledge and difficulty requirements of training and test math problems. Their results show that models can generalize to higher difficulty levels when the required knowledge is the same, but remain weak at chaining compositional skills or adopting novel, unconventional strategies \cite{sun2025omega}. Similarly, e3 finds that only problems with a sufficiently large verification–generation gap benefit from test-time scaling, through experimenting under settings where the problem types of training and test sets are strictly controlled \cite{setlur2025e3}. However, outside such carefully constrained conditions—in typical scenarios where both training and test sets consist of heterogeneous problems with uncontrolled knowledge and difficulty—studies consistently find that RLVR does not substantially expand capability. Theoretical analysis conducted by \citeauthor{wu2025invisibleleash} further argues that, in general, the shrinkage of empirical support outweighs its expansion in such scenarios, leading to little capability gain in RLVR \cite{wu2025invisibleleash}. In this work, we analyze RLVR under such general, uncontrolled math problem settings. By examining how accuracy shifts across difficulty levels, we show that RLVR often fails to improve capability as it tends to deliver gains on easier problems at the expense of performance on harder ones.

\noindent \textbf{Reasoning pattern and knowledge in distillation.} Several studies have examined the respective roles of domain knowledge and reasoning patterns in improving accuracy through distillation. For instance, \citeauthor{shrestha2025warmup} distill teacher responses from logic puzzles—where domain knowledge is minimal—and show that transferring reasoning patterns alone can yield substantial performance gains across domains such as mathematics and coding \cite{shrestha2025warmup}. Similarly, \citeauthor{huan2025mathreasoning} demonstrate that distilling math problem responses leads to notable improvements in reasoning tasks in other domains \cite{huan2025mathreasoning}. However, work on capability remains limited. \citeauthor{yue2025does} suggest that distillation can drive capability expansion, but their analysis does not disentangle the effects of reasoning patterns and knowledge injection \cite{yue2025does}. In contrast, our study explicitly controls for this distinction and investigates how each factor differentially influences model capability.

\section{Accuracy and Capability} \label{sec:section 3}

\subsection{Formal Definitions}
We evaluate models along two dimensions: \textit{accuracy} and \textit{capability}. Informally, accuracy measures how likely a model is to generate a correct answer in a single attempt, while capability measures whether a correct answer exists within the model’s response distribution.

Formally, we define accuracy and capability with respect to given model $M$ and 
evaluation dataset \(\mathcal{D} = \{1, \dots, N\}\) of \(N\) questions. 
Let \(p_i^M\) denote the probability that model \(M\) successfully solves question \(i\) in a single attempt. 
Note that this can be obtained by sampling model \(M\) for \(n\) times on question \(i\), computing the fraction of correct responses, and taking the limit as \(n \to \infty\). In theory, an LLM using softmax sampling assigns non-zero probability to all valid outputs, so any answer could eventually be produced. To make capability practically meaningful, we consider a question \(i\) to be \emph{in-distribution} for model \(M\) if \(p_i^M > \epsilon\), where \(\epsilon\) is a small threshold (typically \(10^{-2}\) to \(10^{-3}\)).

To evaluate performance under multiple attempts, let \(p_{i,k}^M\) denote the probability that model \(M\) solves question \(i\) at least once across \(k\) independent attempts. This probability satisfies 
\[
p_{i,k}^M = 1 - (1 - p_i^M)^k.
\]

With these definitions in place, we define the model's \emph{accuracy} as the average success rate over the entire dataset:
\[
\mathrm{Acc}(M) = \frac{1}{N} \sum_{i \in \mathcal{D}} p_i^M.
\]
We define the model's \emph{pass@k capability} as the average success probability over \(\mathcal{D}\) given \(k\) passes per question:
\[
\mathrm{Cap}_k(M) = \frac{1}{N} \sum_{i \in \mathcal{D}} p_{i,k}^M = \frac{1}{N} \sum_{i \in \mathcal{D}} \left(1 - (1 - p_i^M)^k\right)
\]
It is important to note that if model \(M'\) has higher accuracy than model \(M\) (\(p_i^{M'} > p_i^M\)) for a specific question, then it will also have higher pass@k capability for that question. However, this relationship does not always hold taking into account the entire dataset. In fact, as shown in Appendix~\ref{appendix:acc_cap_example}, it is possible for \(\mathrm{Acc}(M') > \mathrm{Acc}(M)\) while \(\mathrm{Cap}_k(M') < \mathrm{Cap}_k(M)\).

\subsection{Estimating Accuracy and Capability}
\label{estimate}

In practice, it is infeasible to compute the exact accuracy and capability of a model, as this would require a prohibitively large number of samples per question. Instead, we estimate these quantities empirically using a finite number of samples \(k\). Let \(X_{i,k}\) be the number of correct responses among \(k\) samples for question \(i\).

We estimate \emph{accuracy} as:
\[
\mathrm{Acc}(M) \approx \frac{1}{N} \sum_{i \in \mathcal{D}} \frac{X_{i,k}}{k}
\]

We estimate \emph{pass@k capability} as:
\[
\mathrm{Cap}_k(M) \approx \frac{1}{N} \sum_{i \in \mathcal{D}} 1(X_{i,k} > 0)
\]

These estimators are unbiased. Throughout this work, we report results using these estimators, typically with $k=256$. We also consider a question $i$ to be out-of-distribution if $X_{i,256} = 0$, that is, none of the 256 responses to question $i$ are correct. Under this definition, we can say with 95\% confidence that \(p_i^M < 1 - (0.05)^{1/256} \approx 0.012\), that is, question \(i\) is truly out-of-distribution under the threshold \(\epsilon = 0.012\).

\section{Why Doesn't RLVR Improve Capability?} \label{sec:section4}

Prior work has shown across multiple models that RLVR yields substantial gains in accuracy but often fails to improve capability, as measured by pass@$k$ with sufficiently large $k$ \cite{yue2025does}. In this section, we extend this observation and analyze the phenomenon in greater depth. Specifically, we aim to answer: \textit{Why does RLVR raise accuracy while leaving capability unchanged or even degraded?}


\subsection{Capability Analysis}\label{sec:4.1}

We first performed pass@$k$ experiments similar to those by \citeauthor{yue2025does}, confirming that RLVR improves accuracy but not capability (Fig.~\ref{fig:pass@k_graph_section4}). Specifically, we evaluated two base models—Qwen2.5-1.5B-Math~\cite{yang2024qwen25math} and Qwen2.5-3B~\cite{hui2024qwen25}—along with their corresponding RLVR-trained versions. The full training details and pass@$k$ results are provided in Appendix \ref{appendix:1.5B-math} and \ref{appendix:passatk_RLVR}, respectively. All models were trained on the MATH training set, which contains 7,500 questions, and evaluated on the MATH 500 test set with 500 problems. These same datasets are also used for all subsequent experiments in this section.

We report here results on the 1.5B model evaluated on the test set due to space constraints. However, the same pattern holds consistently across both the train and test sets, and across both model sizes (1.5B and 3B). Full results are provided in Appendix \ref{appendix:successrate}. For clarity, we refer to the original 1.5B model as the base model and the RLVR-trained version as the RL model.

\subsection{A Deeper-Dive: Analysis Based on Question Difficulty} \label{section4.2}

\begin{figure}[htbp]
  \centering
  \includegraphics[width=0.46\textwidth]{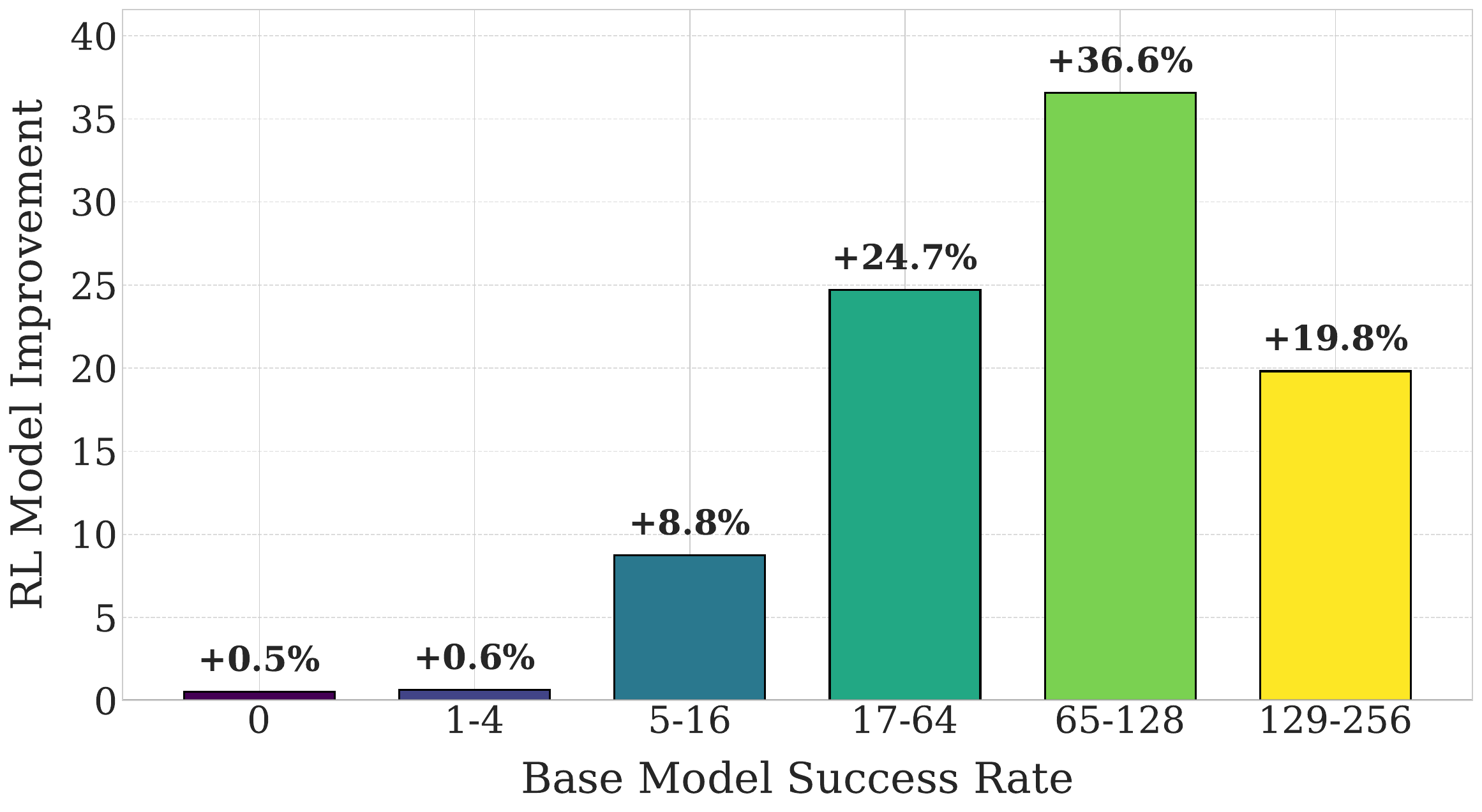}
  \caption{Change in success rates after RLVR training on MATH 500 test set, grouped by the base model's per-question success rate measured over 256 responses. The bar height represents the absolute difference (\%) between the RL and base models within each bin.}
  \label{fig:success_rate_change_bar}
\end{figure}

To better understand the accuracy–capability dynamics of RLVR, we conduct a fine-grained analysis at the question difficulty level. For each question in the training and test sets, we generate 256 responses from the base model and compute its per-question success rate. Questions are then grouped into bins according to these rates: [0], [1–4], [5–16], [17–64], [65–128], and [129–256]. Within each bin, we collect the corresponding questions, retrieve the RL model’s responses to the same questions, and compute average success rates for both models. We then calculate the average success rate of the base and RL models in each bin and plot their absolute difference. The results are shown in Figure~\ref{fig:success_rate_change_bar}.

\begin{figure}[htbp]
  \centering
  \includegraphics[width=0.4\textwidth]{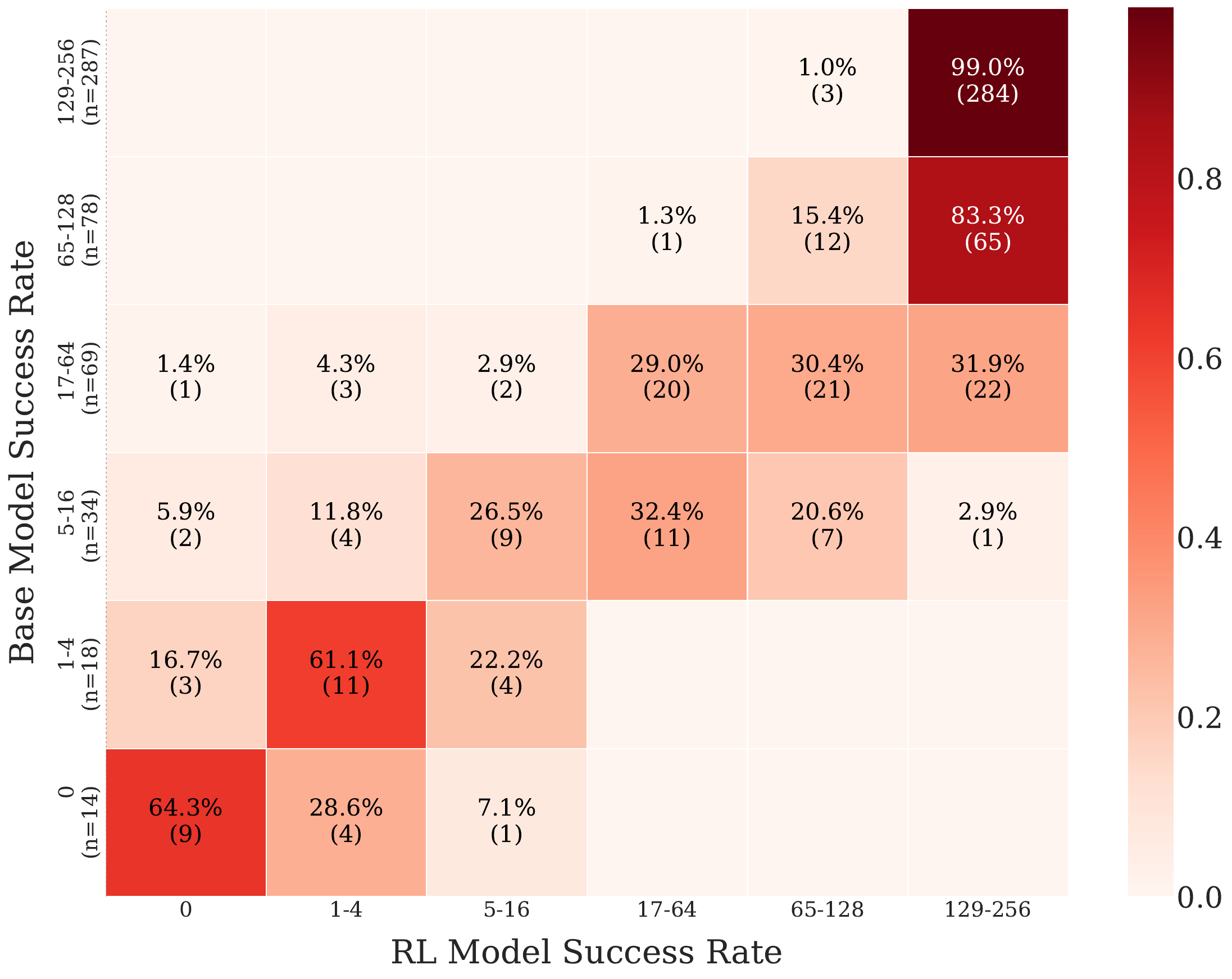}
  \caption{Success rate transition matrix showing redistribution of questions from base model to RL model on MATH 500 test set.}
  \label{fig:heatmap}
\end{figure}

We observe two consistent patterns. (1) For questions where the base model already has a moderately high success rate, the RL model shows substantial improvement. For example, in the training set, the [65–128] bin shows a 36.6\% gain in average success rate relative to 256 responses, and the [17–64] bin shows a 24.7\% gain. (2) In contrast, for questions where the base model has zero or near-zero success rate, the improvement is negligible. For instance, the [0] bin in the training set improves by only 0.5\% points, and the [1–4] bin by just 0.6\% points.

Furthermore, to better understand the pattern observed in success rate improvements, we visualize how individual questions move across success rate bins before and after RLVR training. Figure~\ref{fig:heatmap} presents this transition in test set as a matrix. Here, each row corresponds to a success-rate bin based on the base model’s performance, and each column corresponds to the same bin based on the RL model’s performance. Each cell shows the percentage (and count) of questions that started in a specific base model bin and end up in a particular RL model bin after training. 

Notably, we observe two clear trends.
(1) Questions already in high-success bins tend to stay there or shift upward after RLVR. For example, in the [65–128] bin, 15.4\% remain in place while 83.3\% move to the top [129–256] bin; only 1.3\% (1 question out of 78) drop lower. A similar upward shift appears in the [17–64] bin.
(2) In contrast, questions in low-success bins—especially those near zero—tend to stagnate or regress. In the [1–4] bin, 61.1\% remain and 16.7\% fall to [0]; likewise, in the [5–16] bin, 44.2\% stay or drop lower. This pattern shows a clearer picture of how RLVR fails to help previously unsolved questions and can even increase their number, as many with a small chance of being answered correctly end up never being solved after training.

To understand this behavior, we need to examine the internal dynamics of RLVR training. Take GRPO algorithm as an example. At each iteration, the model generates multiple responses (e.g., 8) per question in a batch. Each response is evaluated for correctness, and parameter updates are applied accordingly. If all 8 responses for a given question are incorrect, that question has no influence on the update. In contrast, for questions with a mix of correct and incorrect responses, the model parameters are nudged to increase the likelihood of generating correct answers and decrease incorrect ones.
Now consider an early training step where the batch includes one difficult question (e.g., in bin [1–4]) and several easier ones (e.g., in bin [65–128]). With high probability, the model will generate only incorrect responses for the difficult question, while producing at least some correct answers for the easier ones. As a result, the parameter update will be guided entirely by the easier questions only. This can lead to the model assigning even lower probability to the correct answers for the difficult question, especially if model capacity is limited.
This selective reinforcement continues throughout training, explaining why questions that initially had a small chance of being answered correctly may become even less likely to be solved after training. To confirm this selective reinforcement effect, we also analyze the entropy of the model's output distributions—measured over 256 responses per question—across difficulty levels. We observe that entropy consistently decreases after RLVR training. Full results are provided in Appendix~\ref{appendix:entropy}.

To summarize, these results suggest the following insight:
RLVR improves accuracy but not capability because \textit{RLVR focuses on improving the accuracy of the less-difficult questions to the detriment of the accuracy of the most difficult questions.}

\section{Comparison of Model Responses Before \& After RLVR}

RLVR causes the model to increase the probability of generating correct answers it could already generate, 
but does not enable it to solve previously unsolvable questions. This naturally leads us to the following question: \textit{Can we replicate RLVR’s effect by directly nudging the model toward its own correct responses—that is, through self-distillation? And if not, 
why?}

\subsection{Self-Distillation Experiments}

\label{sec:self-distillation}

\begin{table}[htbp]
  \centering
  \begin{tabular}{lll}
    \toprule
    \textbf{Setup} & \textbf{Train Acc (\%)} & \textbf{Test Acc (\%)} \\
    \midrule
    1) Base                    & 64.0           & 62.6 \\
    2) Base $\leftarrow$ Base  & 74.7 (+10.7)   & 63.4 (+0.8) \\
    3) RL                      & 80.9           & 74.8 \\
    4) RL \hspace{0.255cm}$\leftarrow$ RL      & 84.4 (+3.5)    & 74.4 (-0.4) \\
    5) Base $\leftarrow$ RL & 80.5 (+16.5) & 74.2 (+11.6) \\
    \bottomrule
  \end{tabular}
  \vspace{0.5em}
  \caption{Self-distillation results for the Qwen2.5-1.5B-Math base model and its RLVR-trained variant. The notation $A$ $\leftarrow$ $B$ indicates the student model $A$ is trained on responses from the teacher model $B$. Values in parentheses show gains over the student model with distillation.}
  \label{tab:self-distillation-1.5b}
\end{table}

To explore this question, we perform self-distillation with rejection sampling, following approaches similar to STaR~\cite{zelikman2022star} and ReST~\cite{gulcehre2023rest}—that is, we conduct SFT on the model’s own correct responses. Recall that, we collected 256 responses per training question from the base model in Section \ref{sec:section4}. From these, we select up to 8 correct responses for each question (using all available correct ones if fewer than 8 exist) and conduct SFT using this filtered dataset. Additionally, to test whether self-distillation can also further improve the RLVR-trained model, we apply the same self-distillation process to the RL model. 
We conduct this experiment for both Qwen2.5-1.5B-Math and Qwen2.5-3B, but present only the 1.5B results here due to space constraints. The 3B results show similar trends and are provided in Appendix~\ref{appendix:self-distillation}.


We find that self-distillation fails to replicate the effect of RLVR. As shown in Table~\ref{tab:self-distillation-1.5b} (lines 1-2), distilling the base model using its own correct responses yields a train accuracy of 74.7\%, a 10.7-point increase. However, test accuracy rises only modestly to 63.4\%, a 0.8-point gain over the base model’s 62.6\% and significantly below the RL model’s 74.8\%. Similarly, from lines 3-4 we observe self-distillation of the RL model leads to no gain on the test set (74.8\% to 74.4\%), despite notable rises in training accuracy (80.9\% to 84.4\%).

These results suggest that, \textit{unlike RLVR, self-distillation tends to overfit to the training set and fails to promote more generalizable reasoning behavior.}

We take a step further and conduct another self-distillation experiment. Prior work has shown that distilling quality responses from a teacher model can effectively improve the performance of student model ~\cite{huang2024o1, min2024imitate, muennighoff2025s1}. In other words, responses that lead to large accuracy gains through distillation can be seen as having high quality ~\cite{kim-etal-2025-evaluating}. 
Based on this idea, we perform an additional experiment: distilling the RL model’s responses into the base model.

Interestingly, the base model shows a significant performance gain that is not limited to the training set. As shown in Table~\ref{tab:self-distillation-1.5b} (line 5), its accuracy on the training set rises from 64.0\% to 80.5\%, as expected after fine-tuning. More importantly, its test accuracy also rises—from 62.6\% to 74.2\%—an absolute gain of 11.6 points, putting it on par with the RL model’s 74.8\%. 

In summary, when distilling the base model with correct responses from the base model there is only minor improvement, whereas when distilling with correct responses from the RLVR-trained model there is significant improvement, on par with RLVR itself.
This suggests that there is a qualitative difference in the two response types, 
and reveals the following insight: RLVR does more than merely increase the success probability for the easier questions—\textit{it enables the model to produce quality responses that were not present in its output distribution before training.}

\subsection{Qualitative Analysis} \label{qualitative_analysis}

\begin{figure}[htbp]
  \centering
  \includegraphics[width=0.48\textwidth]{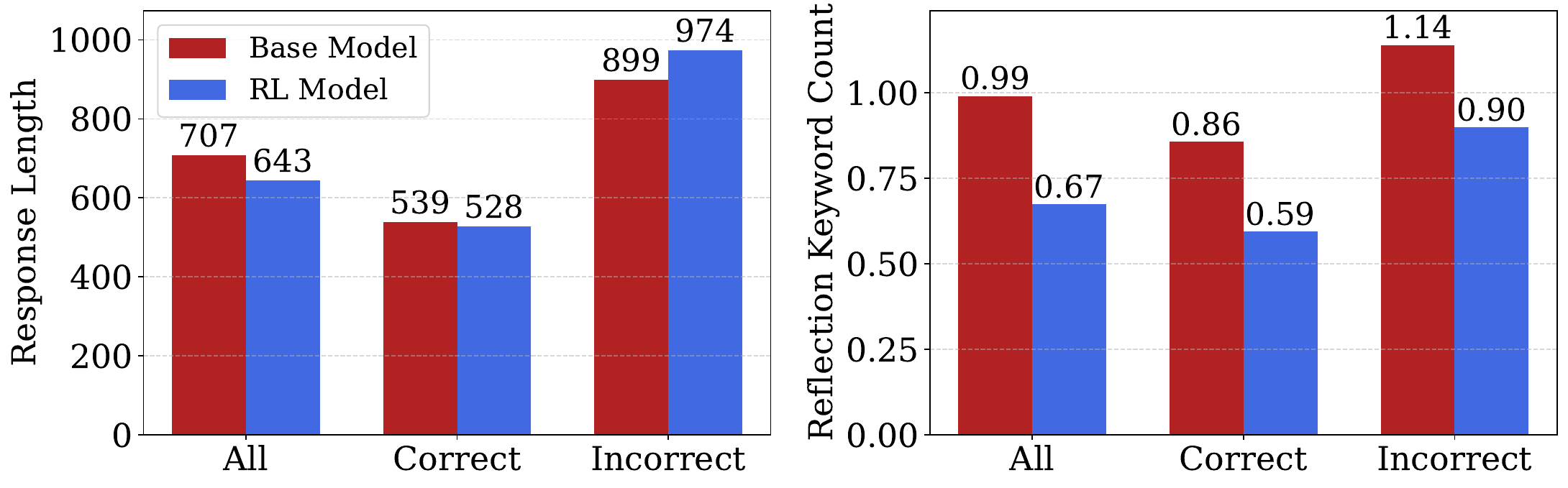}
  \caption{Qualitative analysis of Qwen2.5-1.5B-Math before and after RLVR training. (Left) Mean response length, and (Right) mean count of reflection keywords, grouped by correctness.}

  \label{fig:response_length_reflection_keywords}
\end{figure}

In addition, given that RLVR improves the model’s ability to generate quality responses, 
we examine how RLVR changes the qualitative characteristics of responses. Prior work suggests that RLVR-trained models often produce longer answers and show reflective reasoning behaviors such as verification and backtracking \cite{deepseek2025r1, gandhi2025cognitive, liu2025understanding, yeo2025demystifying, zeng2025simplerl}. Following these,
we compare responses from the base and RL models along two surface-level dimensions: response length and the frequency of reflection-related keywords (e.g., “let’s verify,” “alternatively,” “wait”).

As shown in Figure~\ref{fig:response_length_reflection_keywords}, in our small model settings (1.5B and 3B), we observe no significant difference in response length. Moreover, the RL model tends to produce more direct responses, using fewer reflection-related keywords (Full results in Appendix \ref{appendix:qualitative_analysis}). These patterns diverge from prior observations in the literature. These findings imply that surface-level traits, such as response length or the frequency of reflection keywords, may not reliably indicate response quality. It also underscores the need for developing better quality evaluation criteria in future work.

\section{Under What Conditions Does Distillation Increase Capability?} 
\label{sec:section6}

Distillation from teacher reasoning models is known to be another effective approach for improving accuracy~\cite{  min2024imitate, muennighoff2025s1, ye2025limo}. Here, we take a step further and investigate: \textit{Can teacher distillation also improve capability, and under what conditions does such improvement occur?}

\subsection{Capability Analysis}


\begin{figure}[htbp]
  \centering
  \includegraphics[width=0.48\textwidth]{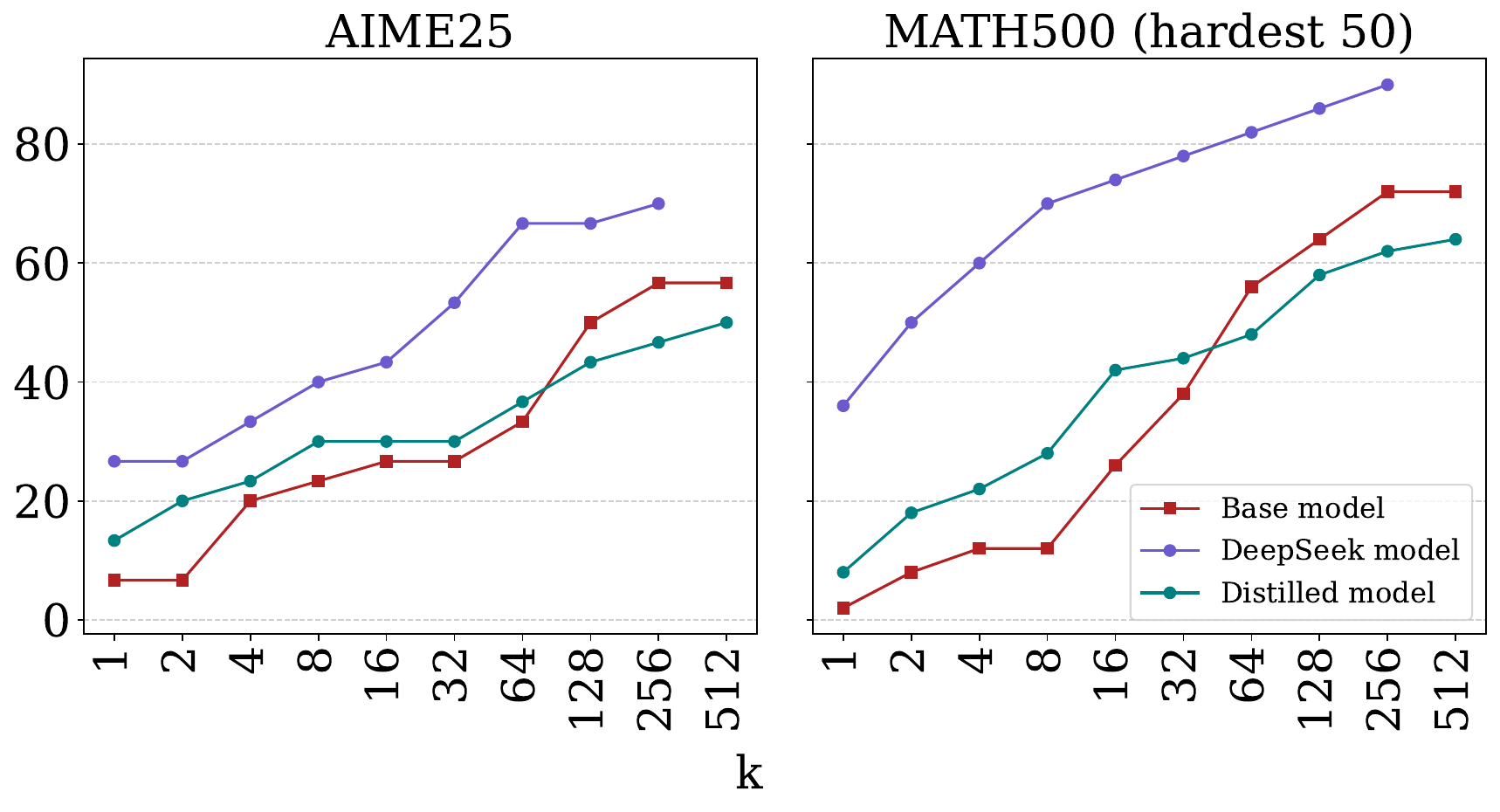}
  \caption{Pass@$k$ comparisons across AIME25 and MATH500 datasets for the 1.5B model and their distillation-trained variants. For the MATH500 results, we show the results on 50 questions with the lowest success rates under the base model to better visualize the performance gaps between models.}
  \label{fig:combined_distillation_pass_at_k}
\end{figure}

\citeauthor{yue2025does} briefly explored this issue by comparing two models: Qwen-2.5-Math-7B~\cite{yang2024qwen25math} and DeepSeek-R1-Distill-Qwen-7B~\cite{deepseek2025r1}. The latter is a publicly available model obtained by distilling 800K DeepSeek-R1 responses into the Qwen-2.5-Math-7B student model. Their experiments showed that the distilled model demonstrates improved capability, as evidenced by higher pass@$k$ scores.

However, it remains unclear why
the distillation process led to the observed capability improvement. Teacher model responses contain two key components: (1) the model’s \textit{reasoning patterns}, and (2) its \textit{domain knowledge}. In the case of DeepSeek-R1-Distill-Qwen-7B, the student model was trained on up to 800K responses generated by DeepSeek-R1~\cite{deepseek2025r1}, making it highly likely that new mathematical knowledge absent in the original student model's pre-training data was also introduced during distillation. As a result, it is difficult to determine whether the observed gains in capability stem from adopting more effective reasoning pattern or from learning new knowledge.

\begin{figure}[htbp]
  \centering
  \includegraphics[width=0.45\textwidth]{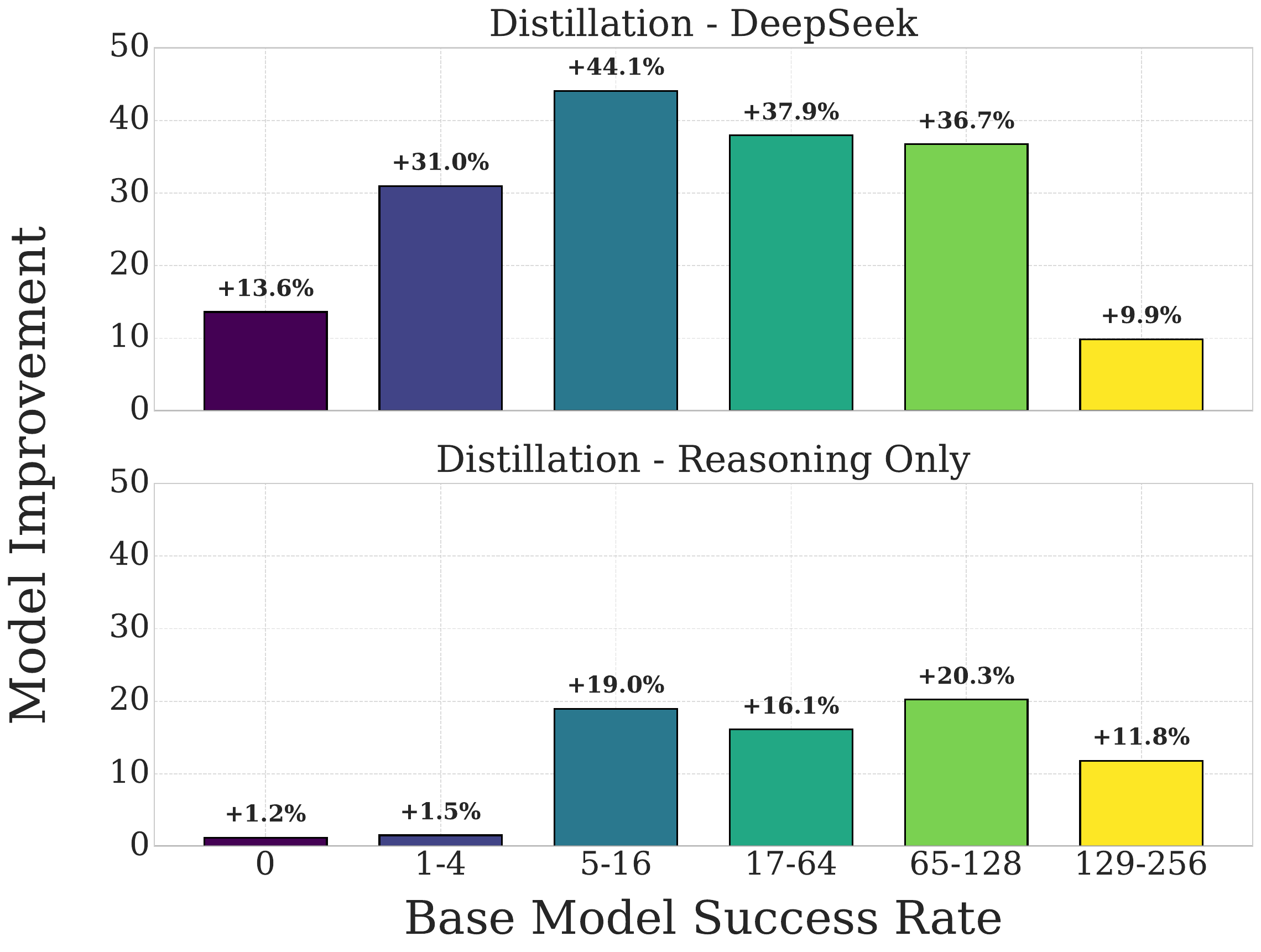}
  \caption{Change in success rates (absolute difference in \%) for the DeepSeek model and the reasoning-only model, grouped by base model success rate bins.}
  \label{fig:success_rate_distilled_combined}
\end{figure}

To disentangle these effects, we conduct a comparative study using three models: (1) Qwen2.5-Math-1.5B, a non-reasoning base model; (2) DeepSeek-R1-Distill-Qwen-1.5B, which, like its 7B counterpart, was trained on 800K responses from DeepSeek-R1~\cite{deepseek2025r1} and likely benefits from both new knowledge and improved reasoning; (3) a distilled model we train ourselves, designed to isolate the effect of reasoning-pattern transfer without introducing any new domain knowledge beyond what the base model already possesses. For simplicity, we refer to these as the base model, the DeepSeek model, and the reasoning-only model.

We train the reasoning-only model using the following procedure. Starting with the base model, we generate 256 responses for each of the 7,500 questions in the MATH train set. We exclude any question for which the model achieves a zero success rate. Since the base model answers each remaining question correctly at least once, we treat these as \textit{in-distribution}. As the teacher, we use QwQ-32B~\cite{qwen2024qwq32b}, which we verified to have higher capability than the student model (see Appendix \ref{appendix:qwq} for details). For each in-distribution question, QwQ-32B generates 8 candidate responses. From these, we randomly select up to 4 correct ones (using all available if fewer than 4 exist) and conduct SFT on the base model with them. Distilling in this manner should not introduce any new knowledge beyond what the base model already has. The resulting distilled model achieves 69\% accuracy on the MATH 500 dataset, significantly outperforming the base model’s 60\%, indicating a successful distillation.

\begin{figure}[htbp]
  \centering
  \includegraphics[width=0.43\textwidth]{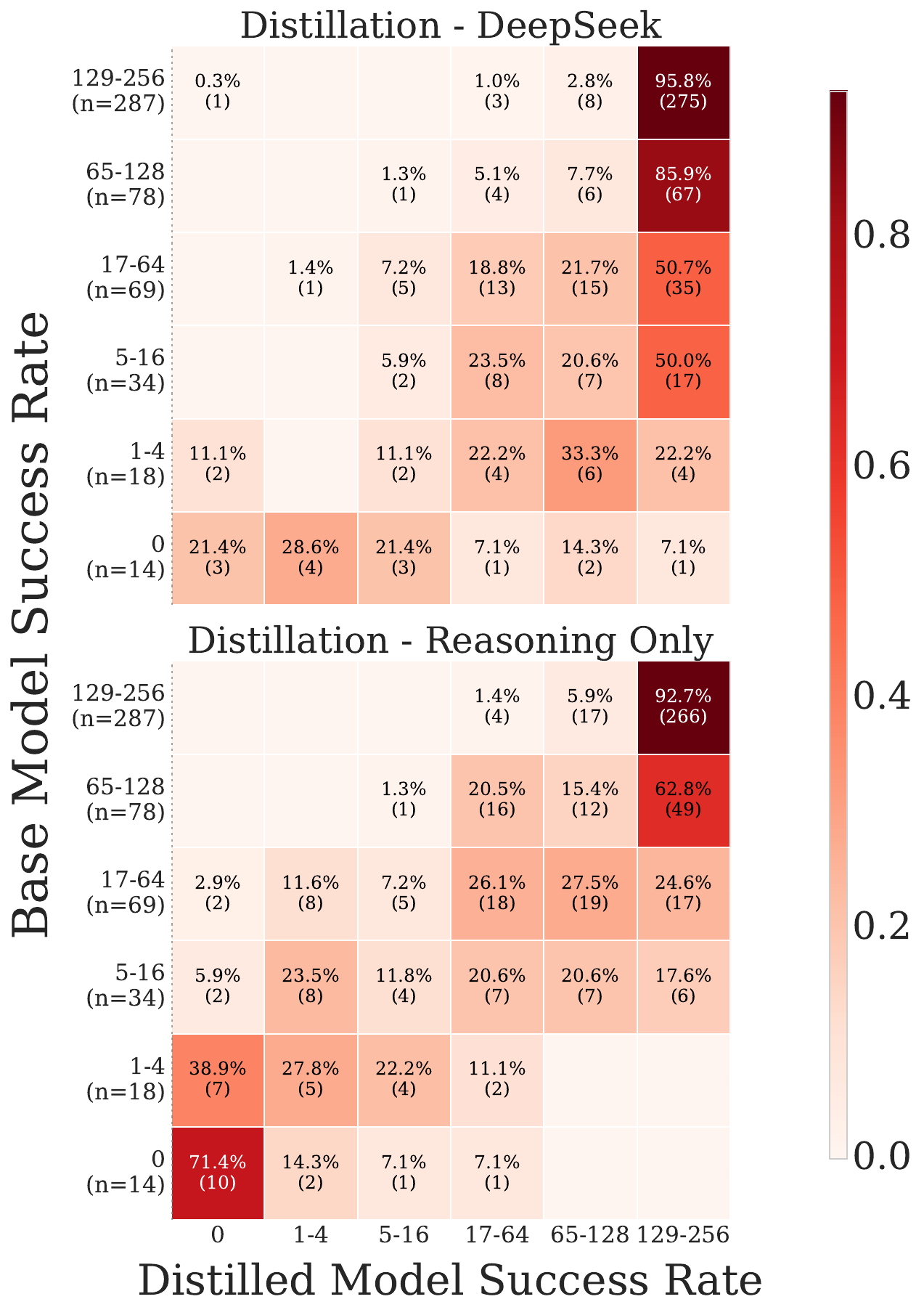}
  \caption{Success rate transition matrix showing redistribution of questions in two distillation settings on MATH 500 test set.}
  \label{fig:heatmap_combined}
\end{figure}

We evaluate the capability of the three models by conducting pass@$k$ experiments on the AIME25 \cite{opencompass_aime2025} and MATH500 dataset. Results are shown in Figure~\ref{fig:combined_distillation_pass_at_k}. We observe the following patterns consistently across both datasets. (1) The DeepSeek model outperforms the base model across the entire range of $k$ values. For AIME25, it achieves a pass@256 score of 70.0\%, compared to 56.7\% for the base model. To ensure this gap does not close at higher $k$, we extend evaluation of the base model up to pass@512 and confirm that its performance plateaus, remaining at 56.7\%. A similar pattern is observed in MATH500. This shows that the DeepSeek model achieves gains in both accuracy and capability, consistent with \citep{yue2025does}. (2) In contrast, the reasoning-only model initially outperforms the base model at lower $k$, but the curves converge and even cross as $k$ increases. We also replicate the experiment with Qwen2.5-3B and observe the same convergence pattern (See Appendix~\ref{appendix:distillation pass@k} for results). These results show that the reasoning-only model we train improves accuracy but does not expand capability, similar to the case of RLVR discussed in Section~\ref{sec:section4}.

These results show that distillation does not always lead to capability expansion, even when it yields significant accuracy gains. We conjecture that the observed difference arises from whether new knowledge is introduced during distillation. Specifically, \textit{Distillation may improve capability when it introduces new knowledge, whereas learning reasoning patterns alone boosts accuracy but not capability.}

\subsection{In-Depth Analysis Based on Question Difficulty} \label{sec:section6.2}

To better understand the dynamics of the two types of distillation, we conduct an analysis similar to that in Section~\ref{section4.2}, comparing model performance across question bins grouped by base model success rates, based on 256 sampled responses. As shown in Figure \ref{fig:success_rate_distilled_combined}, we observe the following patterns: (1) The DeepSeek model shows substantial improvement across all bins, including those with zero or near-zero success rates. (2) In contrast, the reasoning-only model shows improvement primarily in bins with moderately high success rates, but little gain in the zero or near-zero bins, mirroring the behavior of RLVR discussed in Section~\ref{sec:section4}.

As in Section~\ref{section4.2}, we track how per-question success rates change before and after two different distillation settings and visualize the transitions in Figure~\ref{fig:heatmap_combined}. We observe:
(1) For the DeepSeek model, questions consistently move to higher success-rate bins, even those that started in low-success bins. For instance, in the [1–4] bin, only 11.1\% (2 out of 18) drop to the [0] bin, and in the [5–16] bin, no question moves downward.
(2) In contrast, for the reasoning-only model, we interestingly observe the same "sacrificing difficult problems" effect seen in RLVR. In the [1–4] bin, 38.9\% (7 out of 18) drop to the [0] bin, and in the [5–16] bin, 29.4\% (10 out of 34) move to lower bins.


We again conjecture that the key factor underlying this difference is whether new knowledge is introduced during distillation. Specifically, distillation with new knowledge improves both accuracy and capability because \textit{it enables the model to solve questions across all difficulty levels, including the most difficult ones}. In contrast, reasoning-only distillation improves accuracy but not capability because, like RLVR, \textit{it focuses on easier questions—often at the cost of performance on the hardest ones}. We hope this result motivates further empirical study to validate this conjecture and clarify the role of new knowledge in capability expansion.

\section{Conclusions}


Recent prior work has shown that RLVR improves accuracy but often fails to improve capability, whereas distillation from a strong teacher typically improves both. In this paper, we provide explanations for these two phenomena, supported by extensive empirical experiments. In summary, the RLVR training algorithm focuses on improving the accuracy of easier problems while paying less attention to difficult ones; thus, difficult problems on the whole continue to have very low success rates, and the model's overall capability remains largely unchanged.  
Distillation, on the other hand, is typically performed with a strong teacher model that may include knowledge not present in the base model, which can lead to capability expansion.  

\noindent Our specific contributions are as follows:
\begin{enumerate}
    \item We explain why RLVR improves accuracy but not capability by showing that it disproportionately favors easier questions to the detriment of harder ones—often degrading performance for difficult questions.
    \item We demonstrate that RLVR yields higher-quality responses, as evidenced by self-distillation experiments, even though surface-level indicators such as response length or reflection keywords fail to capture this improvement.
    \item While distillation consistently improves accuracy, its effect on capability is less clear. 
    We conjecture that capability improves only when new knowledge is introduced, whereas distilling reasoning patterns only improves accuracy but not capability, sacrificing performance on the most difficult questions, similar to RLVR.
\end{enumerate}
Together, these findings provide a clearer and more nuanced understanding of how RLVR and distillation shape reasoning behavior in LLMs.


\section{Limitations}
While our study provides an in-depth analysis of RLVR and distillation, it also has limitations that suggest directions for future work. 

First, due to resource constraints, our experiments are restricted to a single domain, mathematics, and different patterns may emerge in other tasks. There remains an ongoing debate about whether RLVR truly improves capability. As discussed in Section~\ref{related_work}, some studies argue that RLVR does not enhance capability in general mathematical settings where both training and test sets contain heterogeneous problems with uncontrolled knowledge and difficulty. Others, however, show that RLVR can indeed expand capability when sufficient training compute is available and when training and test sets are carefully controlled in terms of problem type and difficulty. Therefore, follow-up work is needed to unify these perspectives and develop a more comprehensive understanding of the phenomenon. 

Second, our experiments are limited to relatively small models (1.5B and 3B) and a single RL algorithm family (GRPO \& Dr. GRPO). Larger models or different RL algorithms may exhibit different dynamics. A more comprehensive study across model scales and training methods is needed to test the generality of our findings.

Third, our distillation experiments are limited in both scale and control. The DeepSeek model used for comparison is distilled on approximately 800k teacher responses and trained from a different base model, whereas our reasoning-only distillation relies on roughly 30k responses from the same model. In addition, when we extend the setup to include teacher responses to out-of-distribution (OOD) questions, we do not observe measurable capability improvement, possibly due to the small number of OOD samples or limited coverage of new knowledge. Consequently, we cannot conclusively determine whether capability gains depend on the introduction of new knowledge. Future work should validate this conjecture under more controlled settings.


\section{Acknowledgement}
We gratefully acknowledge the support of the Center for AI and Robotics (CAIR) at New York University Abu Dhabi for this research.

\newpage

\bibliography{custom}

\appendix


\newpage 
\onecolumn
\section{Appendix}
\label{sec:appendix}

\subsection{Accuracy vs. Capability Example} \label{appendix:acc_cap_example}
\noindent
As discussed in Section \ref{sec:section 3}, we provide an example to illustrate that a model can have higher \emph{accuracy} but lower \emph{capability} on an evaluation dataset with more than one question.

\noindent Recall the definitions:
\begin{align*}
\mathrm{Acc}(M) &= \frac{1}{N} \sum_{i=1}^N p_i^M, \\
\mathrm{Cap}_k(M) &= \frac{1}{N} \sum_{i=1}^N \left(1 - (1 - p_i^M)^k \right).
\end{align*}

\noindent We compare two models, \( M_1 \) and \( M_2 \), on a toy dataset of \( N=3 \) questions. Their single-attempt success probabilities \( p_i^M \) are shown below:

\begin{table}[h]
\centering
\begin{tabular}{c|ccc}
\toprule
\textbf{Question} & \( p_i^{M_1} \) & \( p_i^{M_2} \) \\
\midrule
1 & 0.9 & 0.5 \\
2 & 0.9 & 0.5 \\
3 & 0.003 & 0.5 \\
\bottomrule
\end{tabular}
\caption{Single-pass success probabilities for models \( M_1 \) and \( M_2 \).}
\end{table}

\noindent We first compute the accuracy of two models on this toy dataset.
\begin{align*}
\mathrm{Acc}(M_1) &= \frac{1}{3}(0.9 + 0.9 + 0.003) = 0.601, \\
\mathrm{Acc}(M_2) &= \frac{1}{3}(0.5 + 0.5 + 0.5) = 0.5.
\end{align*}
\noindent Thus, \( M_1 \) has higher accuracy.

\noindent We now compute capability with \( k = 256 \), which is large enough to expose the low success probability on Question 3 for \( M_1 \):

\noindent Using the formula:
\[
p_{i,k}^M = 1 - (1 - p_i^M)^k,
\]
we compute:

\vspace{0.5em}
\noindent\textbf{Model \( M_1 \):}
\begin{align*}
p_{1,256}^{M_1} &= 1 - (1 - 0.9)^{256} \approx 1, \\
p_{2,256}^{M_1} &= 1 - (1 - 0.9)^{256} \approx 1, \\
p_{3,256}^{M_1} &= 1 - (1 - 0.003)^{256} \approx 0.537. \\
\mathrm{Cap}_{256}(M_1) &= \frac{1}{3}(1 + 1 + 0.537) \approx 0.845.
\end{align*}

\vspace{0.5em}
\noindent\textbf{Model \( M_2 \):}
\begin{align*}
p_{i,256}^{M_2} &= 1 - (1 - 0.5)^{256} = 1 - 2^{-256} \approx 1 \quad \text{for all } i, \\
\mathrm{Cap}_{256}(M_2) &= \frac{1}{3}(1 + 1 + 1) = 1.0.
\end{align*}

\noindent As shown, although \( M_1 \) has significantly higher probabilities to the first two questions—resulting in higher overall accuracy—its probability on the third question is extremely low. As a result, even with many sampling attempts, \( M_1 \) is unlikely to solve all questions. In contrast, \( M_2 \) maintains moderate but consistent success probabilities across all three questions, which leads to a higher chance of solving every question at least once when given sufficient attempts. 

\newpage
\subsection{Pass@$k$ Experiments Results Before \& After RLVR} \label{appendix:passatk_RLVR}

In this paper, we used two models to evaluate the effect of RLVR training: Qwen2.5-1.5B-Math, and Qwen2.5-3B. For corresponding RL model of 1.5B model, we used the Qwen2.5-Math-1.5B-Oat-Zero, a publicly available model trained with MATH train dataset by \citeauthor{liu2025understanding}. For Qwen2.5-3B, we conducted the RLVR training ourselves, also with MATH train dataset. Further details for training can be found at Appendix \ref{appendix:1.5B-math} and \ref{appendix:3b}, respectively.

\begin{table}[ht]
\centering
\begin{tabular}{llccc@{\hskip 1em}ccc}
\toprule
\multirow{2}{*}{\textbf{Split}} & \multirow{2}{*}{\textbf{Model}} 
& \multicolumn{3}{c}{\textbf{Qwen2.5-1.5B-Math}} 
& \multicolumn{3}{c}{\textbf{Qwen2.5-3B}} \\
\cmidrule(lr){3-5} \cmidrule(lr){6-8}
& & Accuracy & Maj@256 & Pass@256 & Accuracy & Maj@256 & Pass@256 \\
\midrule
\multirow{3}{*}{Train} 
  & Base       & 64.0\% &  76.8\% &  97.2\% & 59.3\% & 80.9\% & 92.7\% \\
  & RL         & 80.9\% &  82.1\% &  97.1\% & 67.9\% & 82.2\% & 92.1\% \\
  & Difference & +16.9\% & +5.3\% &  -0.1\% & +8.6\% & +1.3\% & -0.6\% \\
\midrule
\multirow{3}{*}{Test} 
  & Base       & 60.6\% & 72.0\%  & 97.2\%  & 54.9\% & 76.5\% & 95.8\% \\ 
  & RL         & 74.2\% & 80.8\%  & 97.0\%  & 63.6\% & 79.5\% & 95.8\% \\
  & Difference & +13.9\% & +8.8\% & -0.2\%  & +8.7\% & +3.0\% & +0.0\% \\
\bottomrule
\end{tabular}
\caption{Performance comparison of base and RL models for Qwen2.5-1.5B-Math and Qwen2.5-3B}
\label{tab:train_test_metrics}
\end{table}

\begin{figure}[htbp]
  \centering
  \includegraphics[width=1\textwidth]{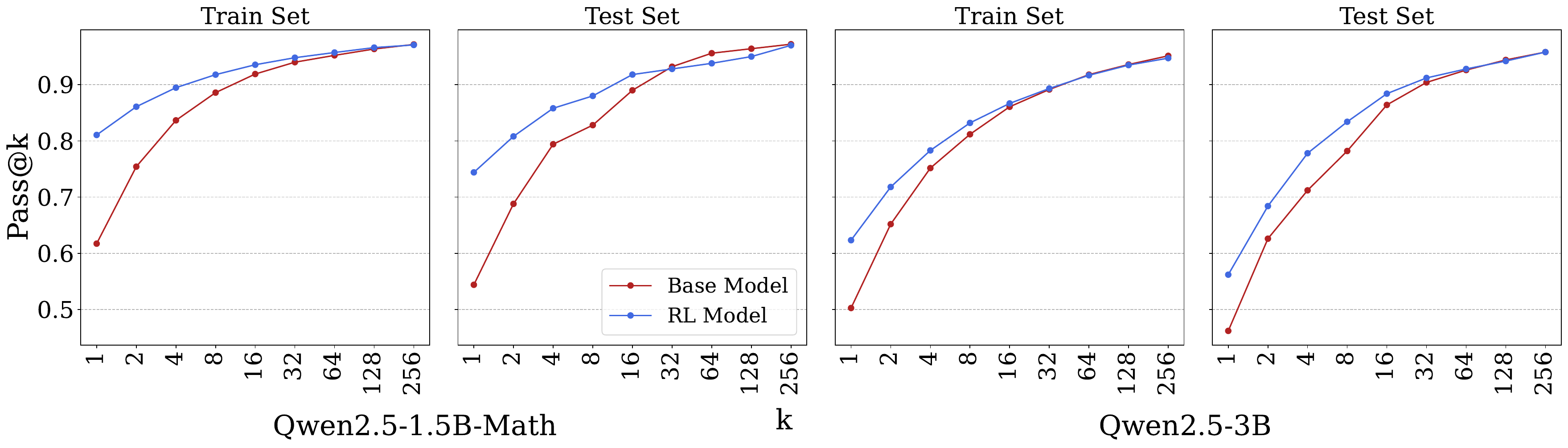}
  \caption{Pass@k comparison between base and RLVR-trained models on train and test sets.}
  \label{fig:pass@k_graph_section4}
\end{figure}

Similar to the work done by \citeauthor{yue2025does}, we conducted the pass@$k$ experiments with these models. For both the base and RL models, we generated 256 responses per question on the MATH train set and MATH500 test set. Using these responses, we estimated accuracy and pass@$k$ capability for $k = 1$ to 256, following the metric defined in Section~\ref{estimate}. Additionally, we computed majority vote accuracy (maj@256), which is the percentage of questions where the most frequent answer among the 256 responses is correct.

As expected, we observed that RLVR significantly improved both accuracy and majority vote performance across training and test sets. As shown in Table \ref{tab:train_test_metrics}, these gains appeared consistently in both the 3B and 1.5B models, indicating that RLVR leads to generalizable improvement in accuracy without signs of overfitting. In contrast, we observed no meaningful improvement in capability. For both the 1.5B and the 3B models, pass@k either remained stable or slightly declined across the training and test sets. As shown in Figure~\ref{fig:pass@k_graph_section4}, the RL model outperformed the base model at small $k$, but their curves converged as $k$ increases—a pattern consistent with prior work~\cite{shao2024deepseekmath, yue2025does}.

\newpage
\subsection{Question-Difficulty-Based Analysis Results} \label{appendix:successrate}

\begin{figure}[htbp]
  \centering
  \begin{subfigure}[t]{0.48\textwidth}
    \centering
    \includegraphics[width=\textwidth]{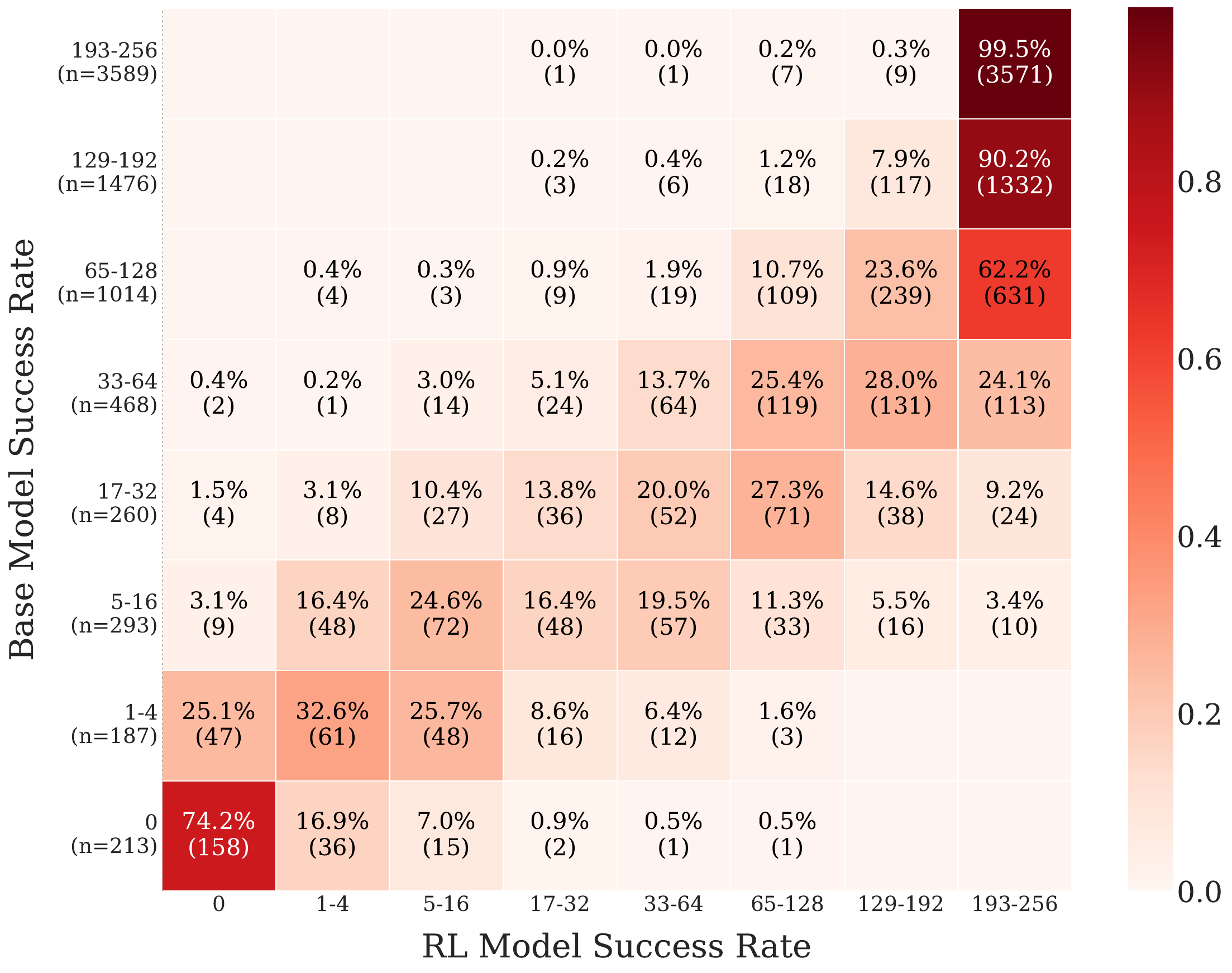}
    \caption{Qwen2.5-1.5B-Math (Train)}
    \label{fig:heatmap_train_1.5b}
  \end{subfigure}
  \hfill
  \begin{subfigure}[t]{0.48\textwidth}
    \centering
    \includegraphics[width=\textwidth]{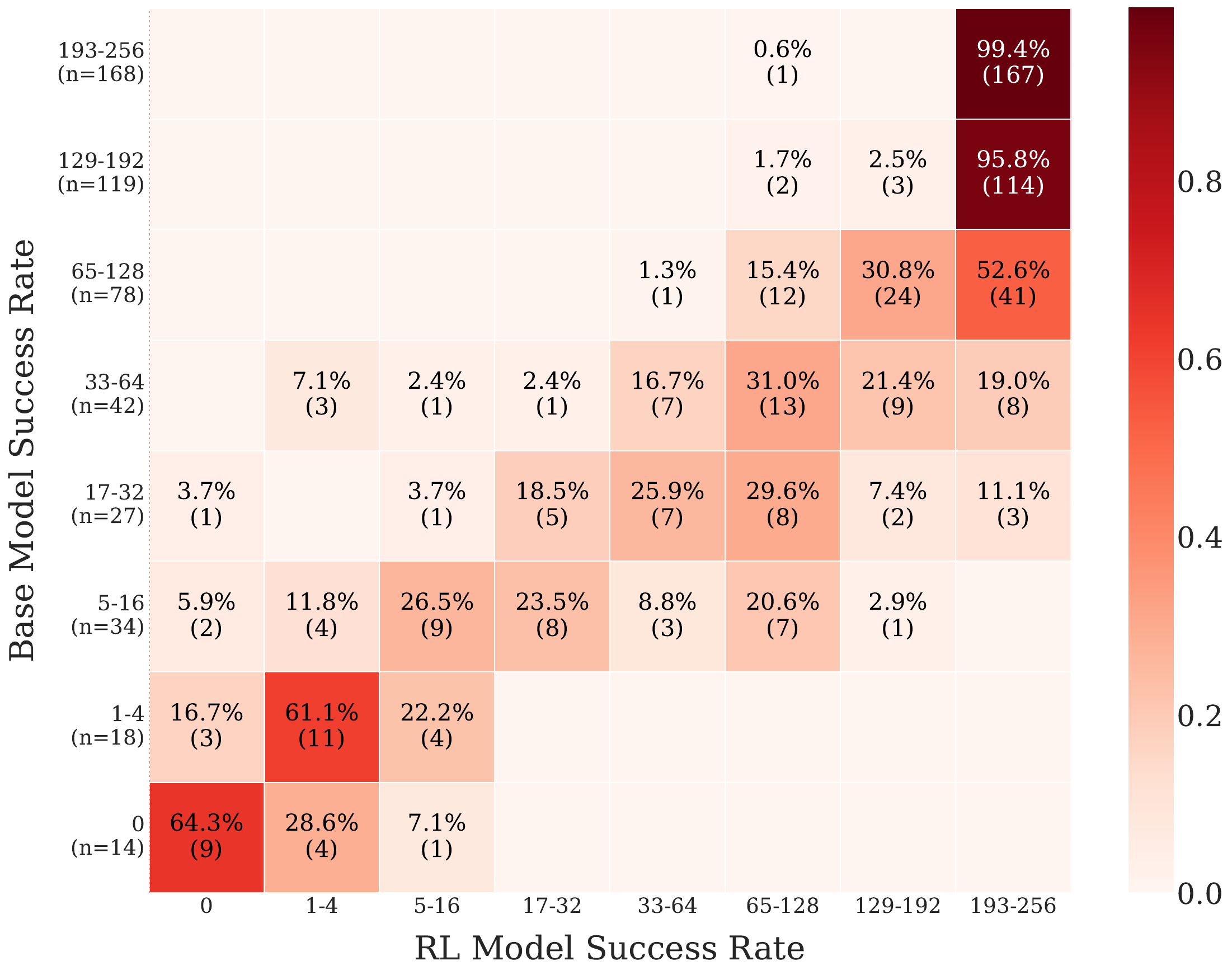}
    \caption{Qwen2.5-1.5B-Math (Test)}
    \label{fig:heatmap_test_1.5b}
  \end{subfigure}

  \vspace{1em}

  \begin{subfigure}[t]{0.48\textwidth}
    \centering
    \includegraphics[width=\textwidth]{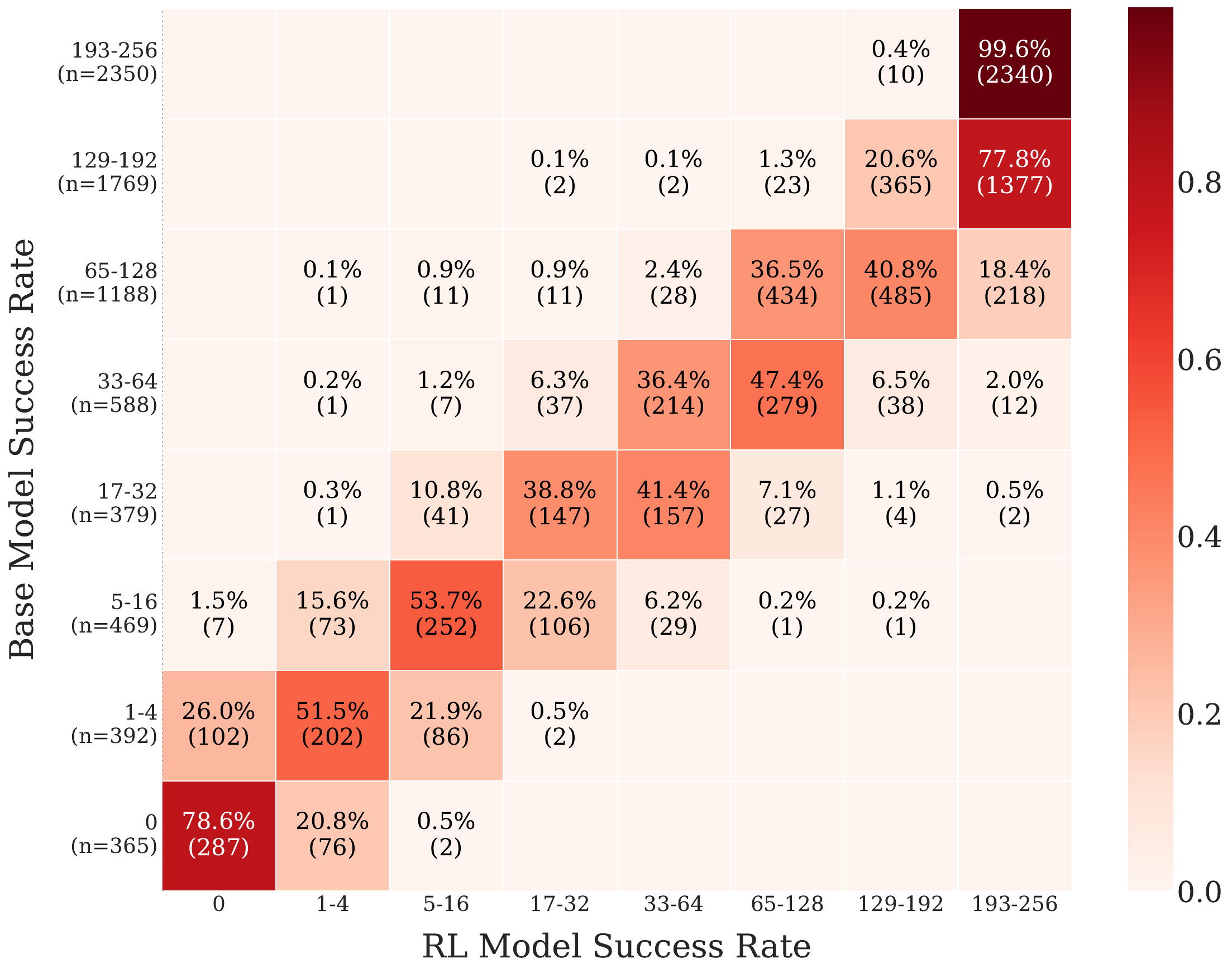}
    \caption{Qwen2.5-3B (Train)}
    \label{fig:heatmap_train_3b}
  \end{subfigure}
  \hfill
  \begin{subfigure}[t]{0.48\textwidth}
    \centering
    \includegraphics[width=\textwidth]{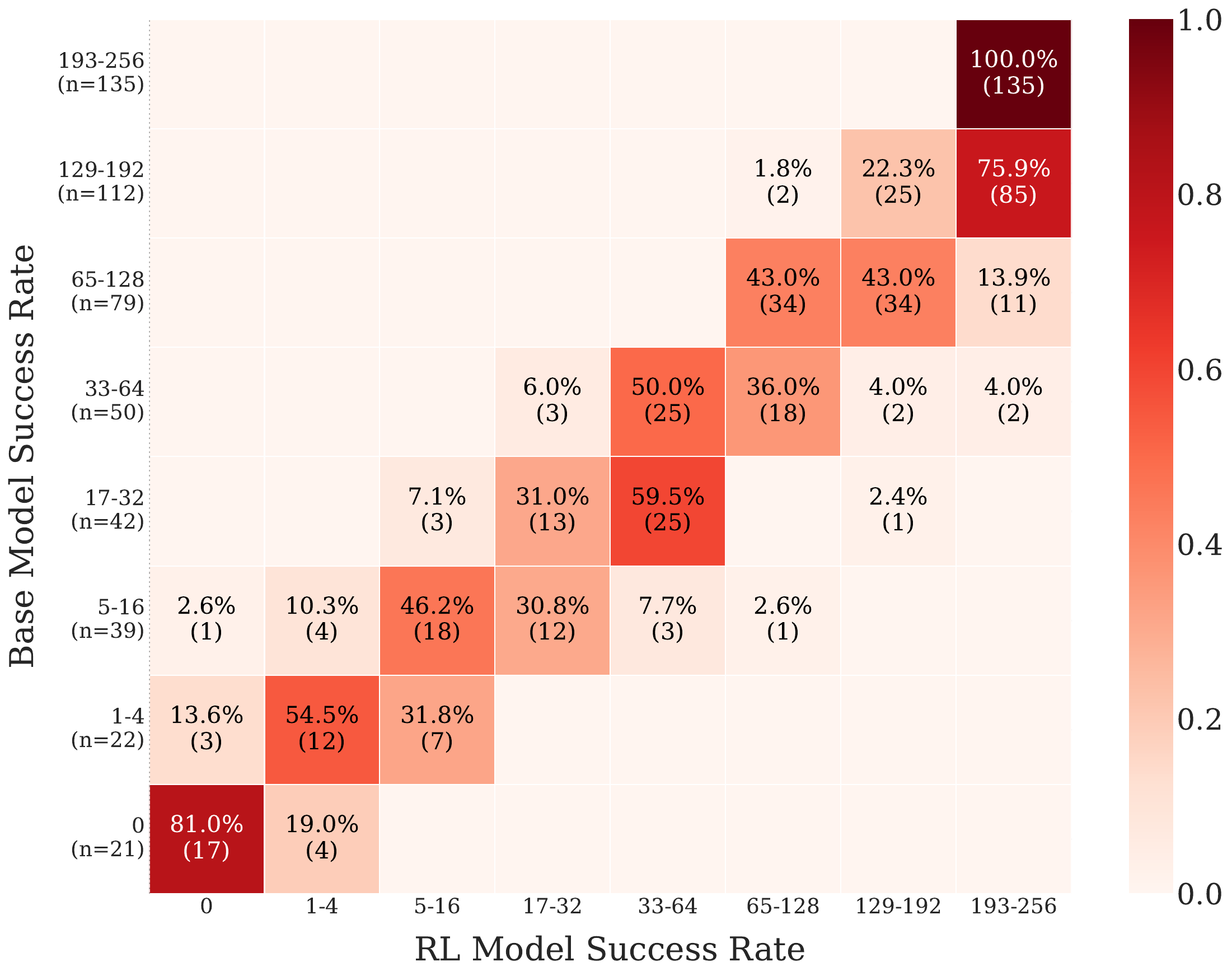}
    \caption{Qwen2.5-3B (Test)}
    \label{fig:heatmap_test_3b}
  \end{subfigure}

  \caption{
Transition matrices comparing base and RLVR success-rate bins for Qwen2.5-1.5B-Math and Qwen2.5-3B. Each cell shows the percentage and count of questions moving between success bins. Most upward transitions occur from mid-success bins; questions in low-success bins are more likely to remain unchanged or regress.
  }
  \label{fig:appendix_heatmap_combined}
\end{figure}

As discussed in Sections~\ref{sec:section4}, we performed detailed analyses based on question difficulty across different training settings. The results are presented below. Figure~\ref{fig:appendix_successrate_combined} shows success rate improvements across difficulty bins for both 1.5B and 3B models on train and test. Figure~\ref{fig:appendix_heatmap_combined} presents the corresponding transition matrices that illustrate how questions move between success rate bins before and after training.

\begin{figure}[htbp]
  \centering
  \begin{subfigure}[t]{0.48\textwidth}
    \centering
    \includegraphics[width=\textwidth]{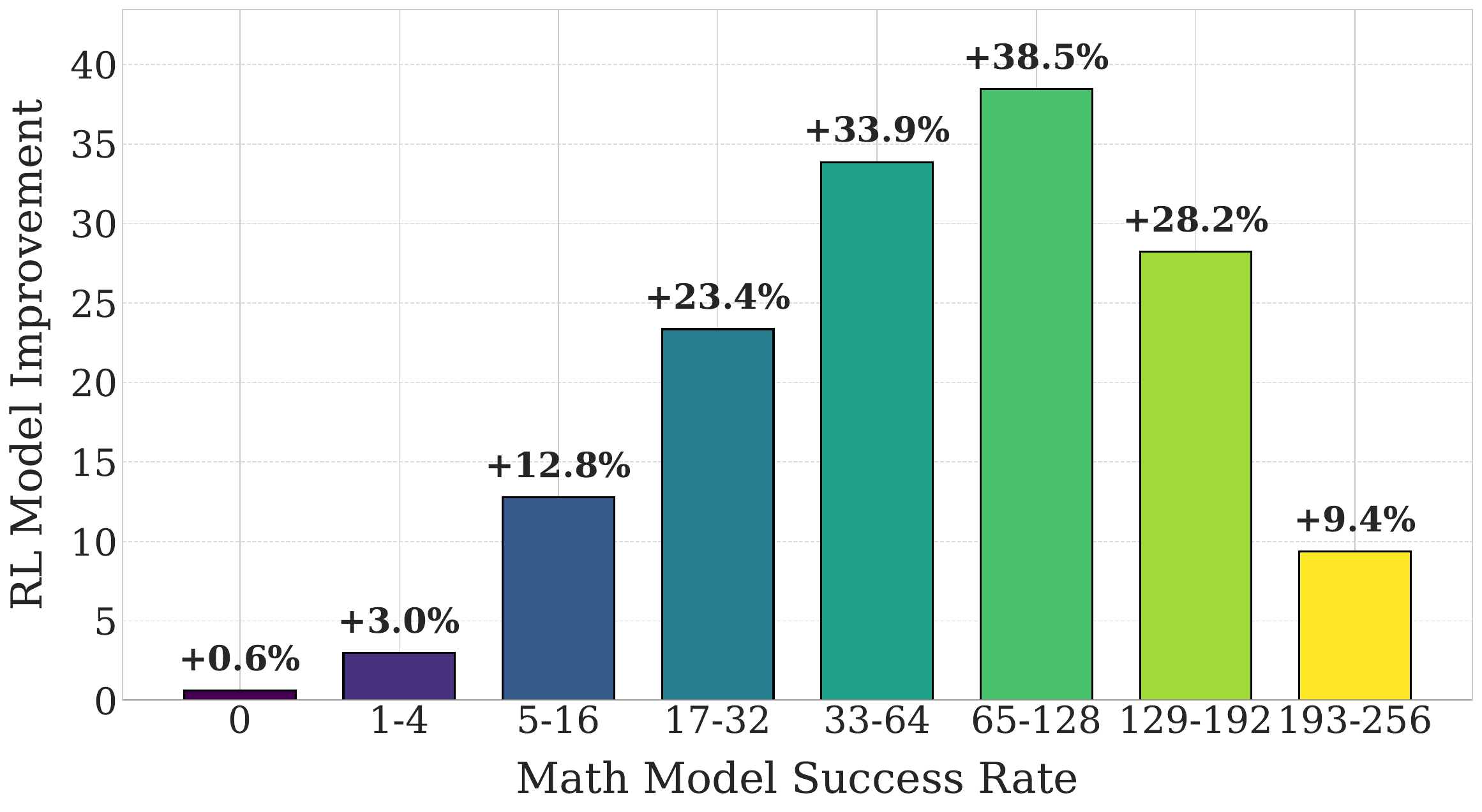}
    \caption{Qwen2.5-1.5B-Math (Train)}
    \label{fig:success_rate_train_1.5b}
  \end{subfigure}
  \hfill
  \begin{subfigure}[t]{0.48\textwidth}
    \centering
    \includegraphics[width=\textwidth]{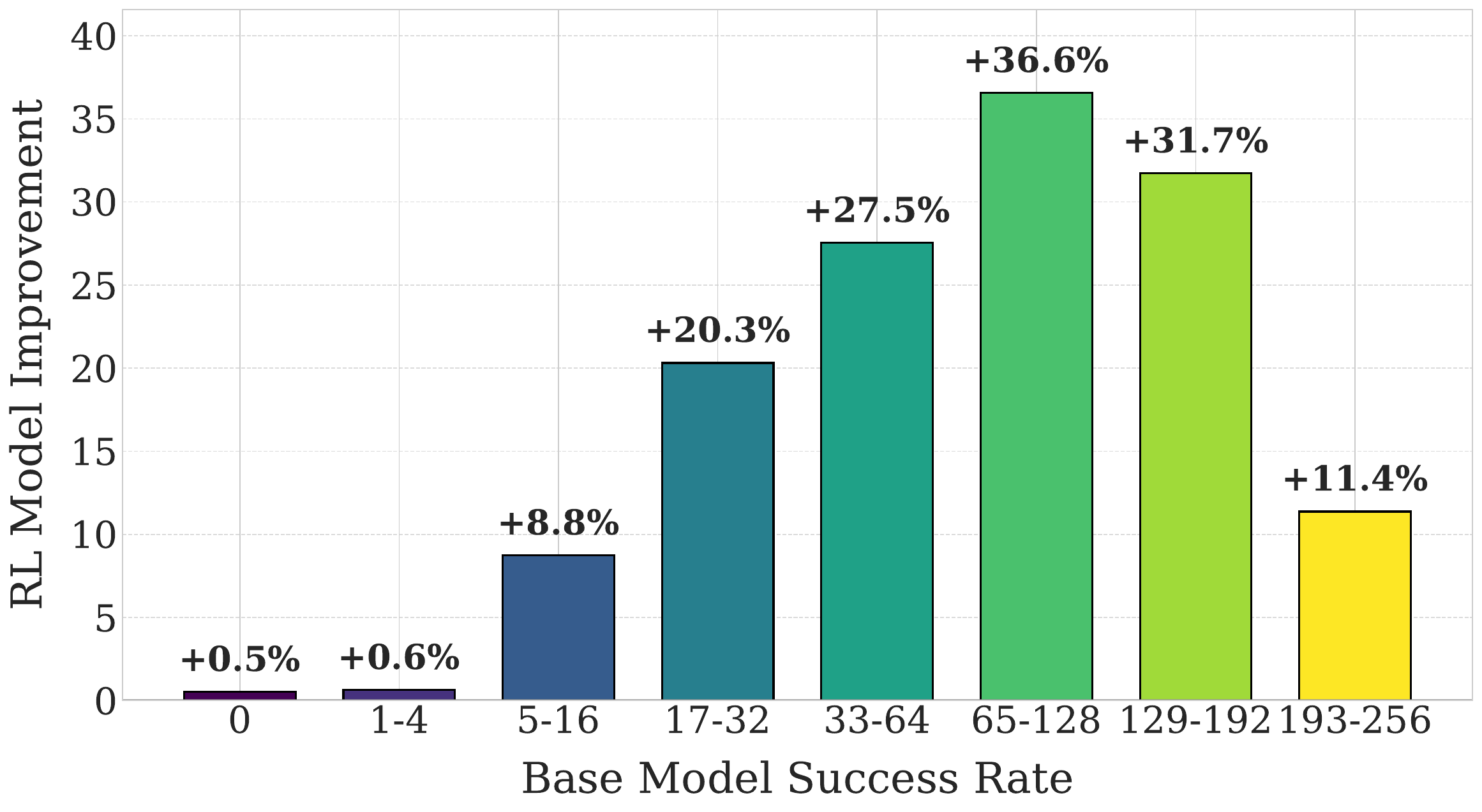}
    \caption{Qwen2.5-1.5B-Math (Test)}
    \label{fig:success_rate_test_1.5b}
  \end{subfigure}

  \vspace{1em}

  \begin{subfigure}[t]{0.48\textwidth}
    \centering
    \includegraphics[width=\textwidth]{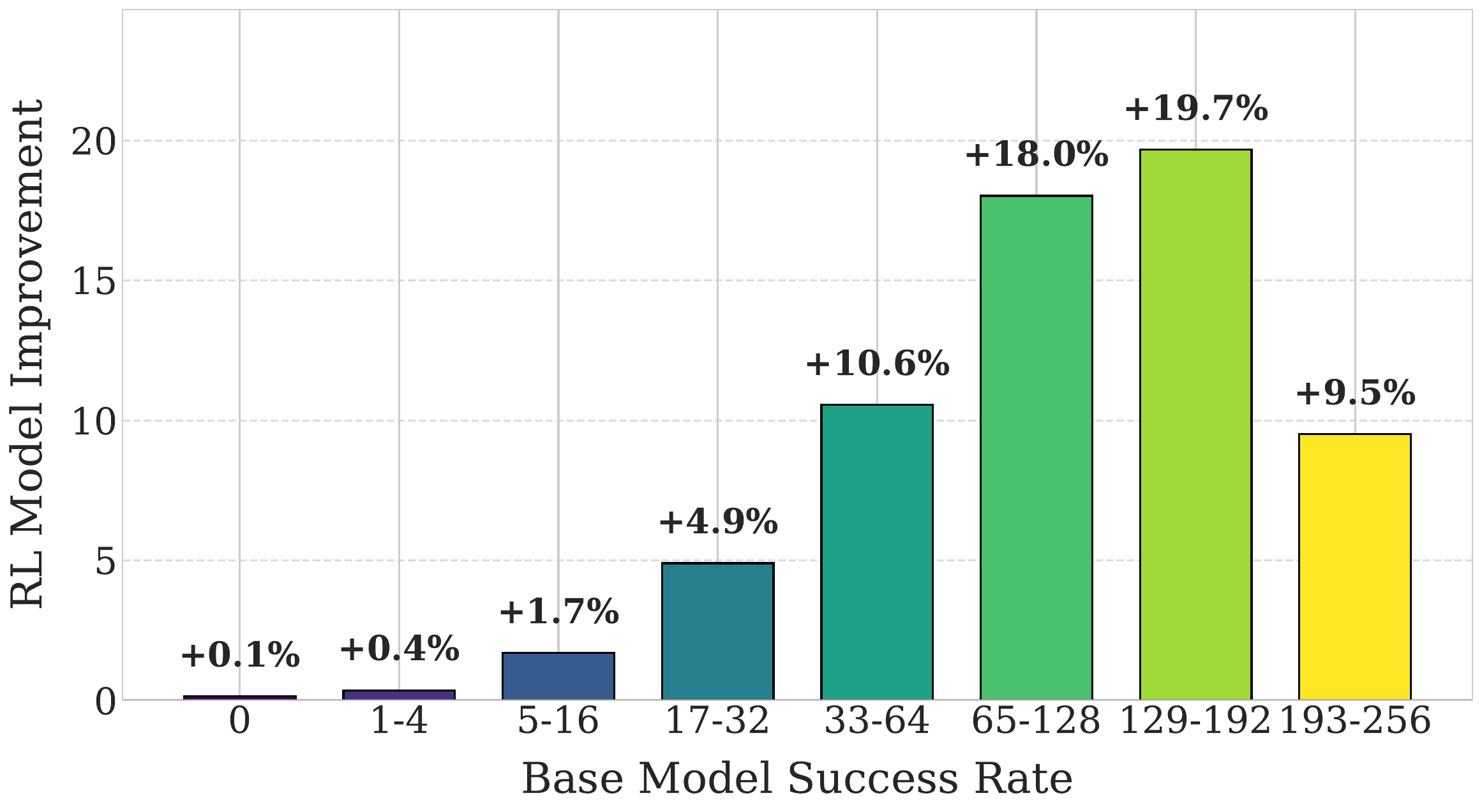}
    \caption{Qwen2.5-3B (Train)}
    \label{fig:success_rate_train_3b}
  \end{subfigure}
  \hfill
  \begin{subfigure}[t]{0.48\textwidth}
    \centering
    \includegraphics[width=\textwidth]{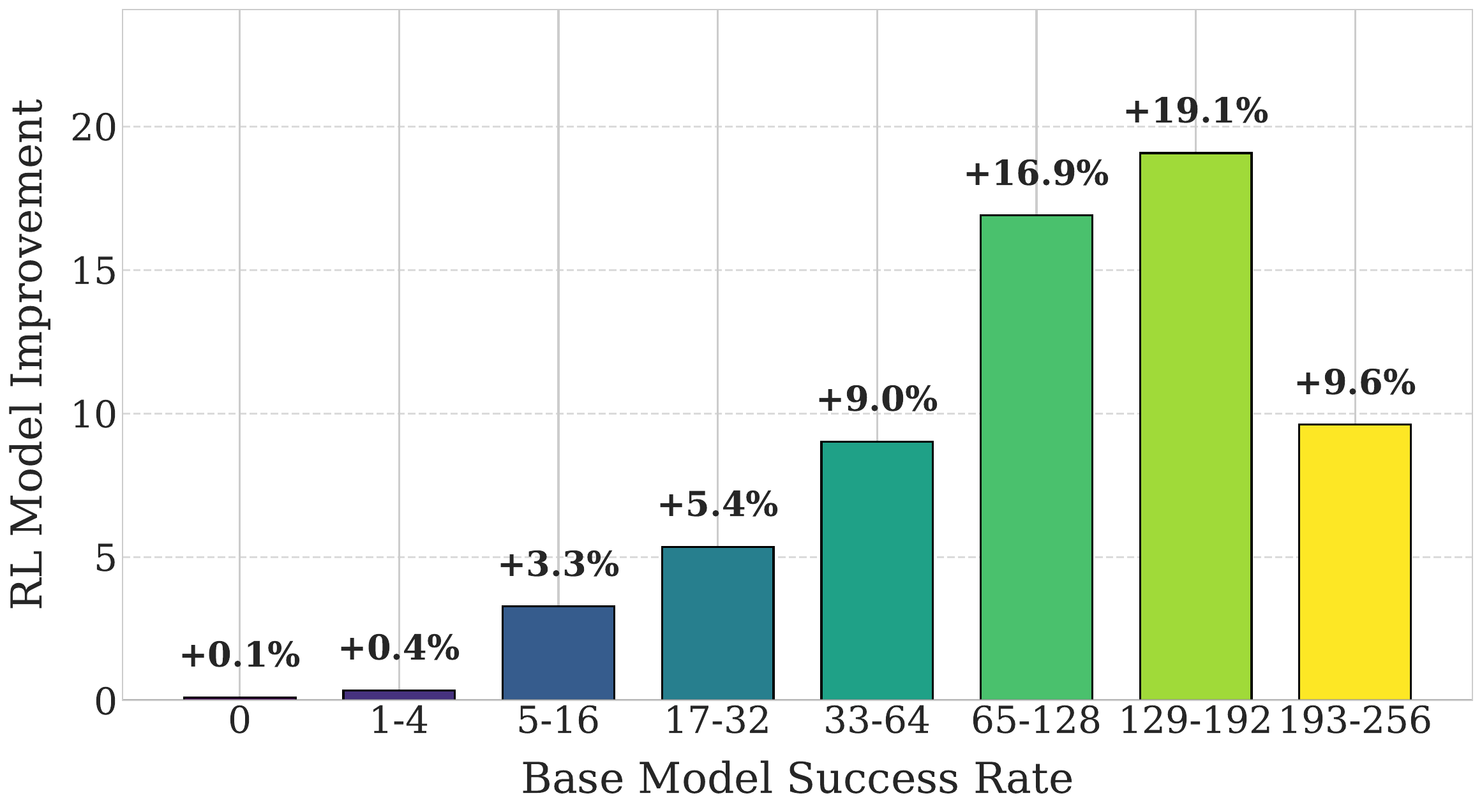}
    \caption{Qwen2.5-3B (Test)}
    \label{fig:success_rate_test_3b}
  \end{subfigure}

  \caption{
Change in success rates (absolute \%) across difficulty bins for Qwen2.5-1.5B-Math and Qwen2.5-3B on the MATH training and test sets. In both models, RLVR significantly improves questions in the mid-success bins (e.g., [17–64], [65–128]), but yields minimal gains in the lowest bins ([0], [1–4]).
  }
  \label{fig:appendix_successrate_combined}
\end{figure}

\newpage
\subsection{Entropy Analysis} \label{appendix:entropy}

\begin{figure}[htbp]
  \centering
  \begin{subfigure}[t]{0.48\textwidth}
    \centering
    \includegraphics[width=\textwidth]{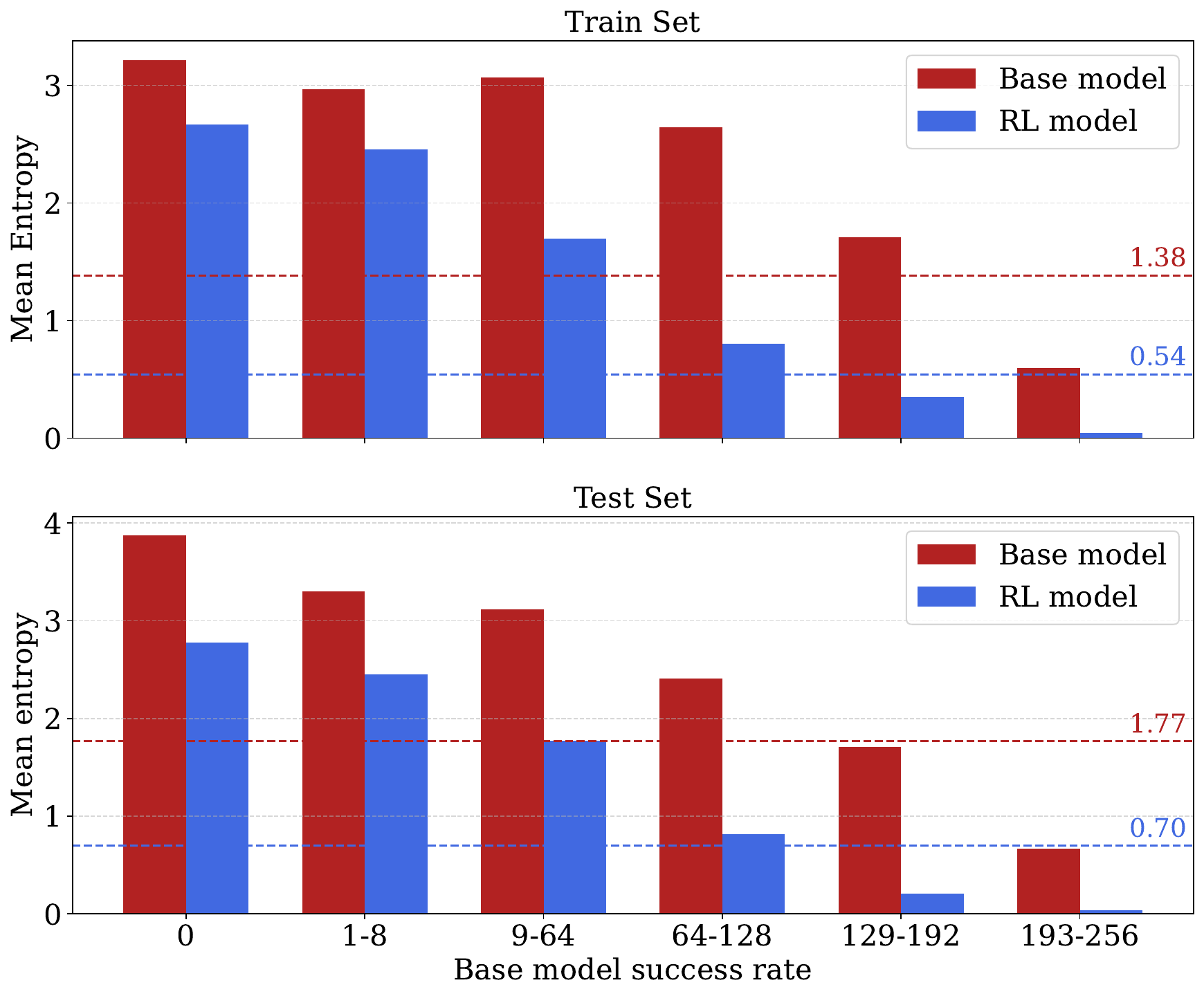}
    \caption{Qwen2.5-1.5B-Math}
    \label{fig:entropy_graph_1.5B}
  \end{subfigure}
  \hfill
  \begin{subfigure}[t]{0.48\textwidth}
    \centering
    \includegraphics[width=\textwidth]{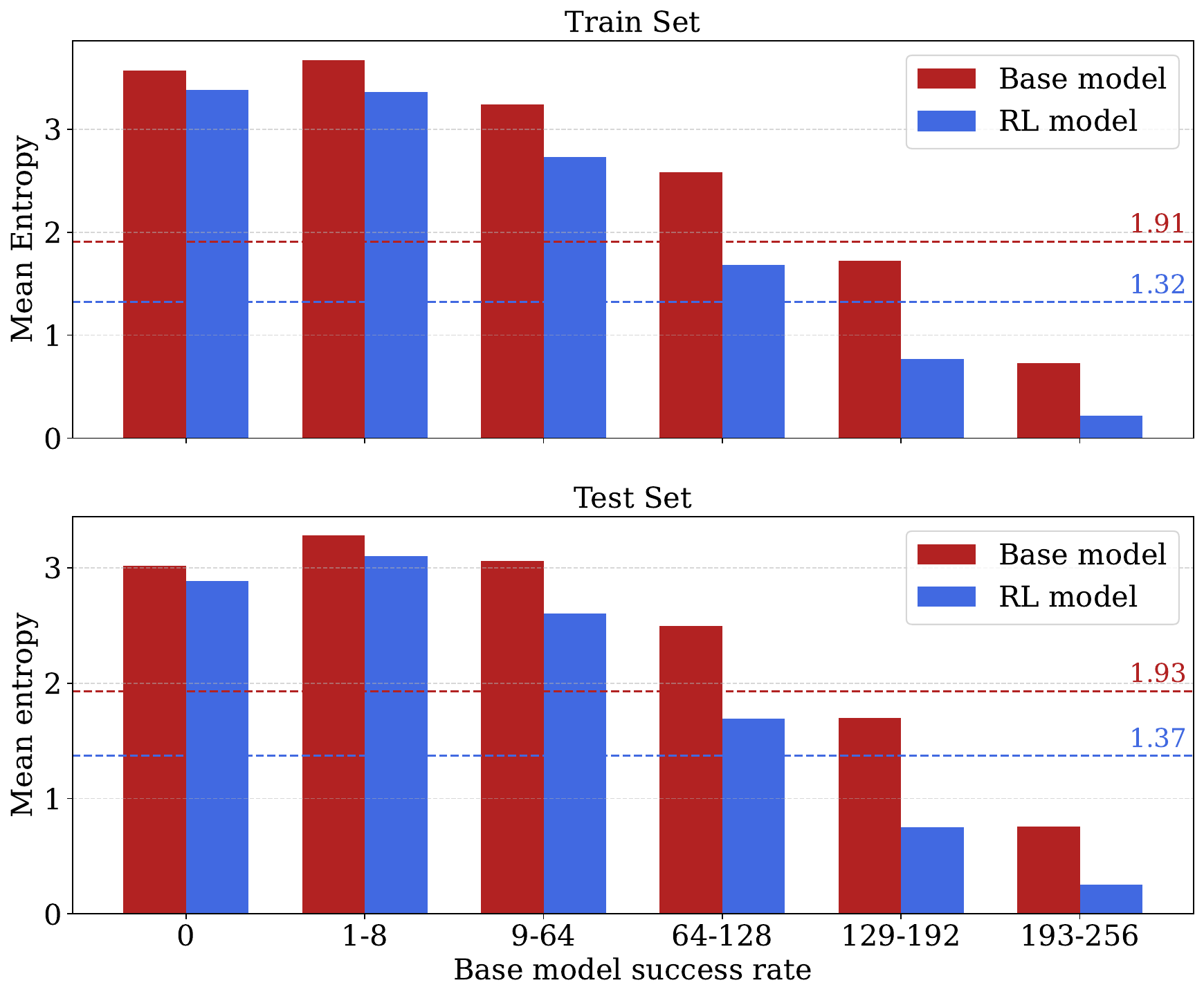}
    \caption{Qwen2.5-3B}
    \label{fig:entropy_graph_3B}
  \end{subfigure}
  \caption{Comparison of output entropy across base and RLVR-trained models, computed over 256 responses per question. Entropy is measured by grouping responses with identical final answers.}
  \label{fig:entropy_comparison}
\end{figure}

As discussed in Section~\ref{section4.2}, we performed an entropy analysis to examine whether RLVR reduces response diversity and causes the model to concentrated on fewer outputs. For both the base and RL models, we computed the entropy of the answers in the 256 responses for each question 
and then average the entropy values over the questions.
Entropy values were measured by grouping responses that yield the same final answer, regardless of correctness. We conducted the experiment for both Qwen2.5-1.5B-Math and Qwen2.5-3B.

As shown in Figure~\ref{fig:entropy_comparison}, output entropy dropped noticeably after RLVR. More importantly, this reduction consistently appeared across all difficulty levels, including questions where the base model had zero or near-zero success rate. These results support the hypothesis that RLVR reinforces greater 
consistency, but on harder questions, this often leads the model to focus on incorrect responses—making it less likely to recover even occasional correct answers.

\newpage

\subsection{Self-Distillation Results} \label{appendix:self-distillation}
Table \ref{tab:self-distillation-appendix} reports the full accuracy results for the self-distillation experiments introduced in Section~\ref{sec:self-distillation}. 
All fine-tuning used the same hyperparameter configuration (Appendix~\ref{appendix:sft_hyperparams}).

\begin{table*}[htbp]
  \centering
  \begin{tabular}{lllll}
    \toprule
    \textbf{Model} & \textbf{Student Model} & \textbf{Teacher Model} & \textbf{Train Accuracy} & \textbf{Test Accuracy} \\
    \midrule
     & Base model &            &   64.0\%   &    62.6\%    \\
     & Base model & Base model &   74.7\% (+10.7\%)         &    63.4\% (+0.8\%)           \\
     Qwen2.5-1.5B-Math & RL model   &      &  80.9\%  &      74.8\%     \\
     & RL model   & RL model   &   84.4\% (+3.5\%)             &    74.4\% (-0.4\%)          \\  
     & Base model & RL model &  80.5\% (+16.5\%)             & 74.2\% (+11.6\%)       \\
       \midrule
    & Base model &            & 59.3\%           & 54.9\% \\
     & Base model & Base model & 73.6\% (+14.3\%) & 58.7\% (+3.8\%) \\
    Qwen2.5-3B & RL model   &            & 67.9\%           & 63.6\% \\
     & RL model   & RL model   & 72.1\% (+4.2\%)  & 64.4\% (+0.8\%) \\
     & Base model & RL model & 73.6\% (+14.3\%) & 64.5\% (+9.6\%) \\

    \bottomrule
  \end{tabular}
  \vspace{0.5em}
  \caption{Self-distillation results for Qwen2.5-1.5B-Math and Qwen2.5-3B across different student–teacher configurations. Accuracy values in parentheses reflect improvements over the corresponding student model before fine-tuning. }
  \label{tab:self-distillation-appendix}
\end{table*}

As discussed in Section~\ref{sec:self-distillation}, the 1.5B model shows that self-distillation leads to overfitting: while training accuracy increases significantly, test accuracy improves only slightly or even declines. In contrast, distilling RL responses into the base model yields the strongest generalization improvement, raising test accuracy from 62.6\% to 74.2\%—surpassing both the base and RLVR-trained models. Similar trends are observed for the Qwen2.5-3B model as well: distilling RL responses into the base model again leads to the highest test accuracy, outperforming all other configurations. This consistent pattern across model sizes reinforces the interpretation that RLVR produces responses of higher quality. Taken together, these results suggest that distillation performance itself may serve as a useful proxy for evaluating the quality of model responses—potentially offering a more meaningful signal than surface-level indicators such as response length or syntactic heuristics.


\subsection{Qualitative Analysis of Responses Before \& After RLVR}
\label{appendix:qualitative_analysis}

In Section~\ref{qualitative_analysis}, we compared responses from Qwen2.5-1.5B-Math and Qwen2.5-3B before and after RLVR training along two dimensions: response length and the use of reflection-related keywords (e.g., "let's verify", "alternatively", "wait"). Here, we present the complete results.

\subsubsection{Response Length}

\begin{figure}[H]
  \centering
  \begin{subfigure}[t]{0.3\textwidth}
    \centering
    \includegraphics[width=\linewidth]{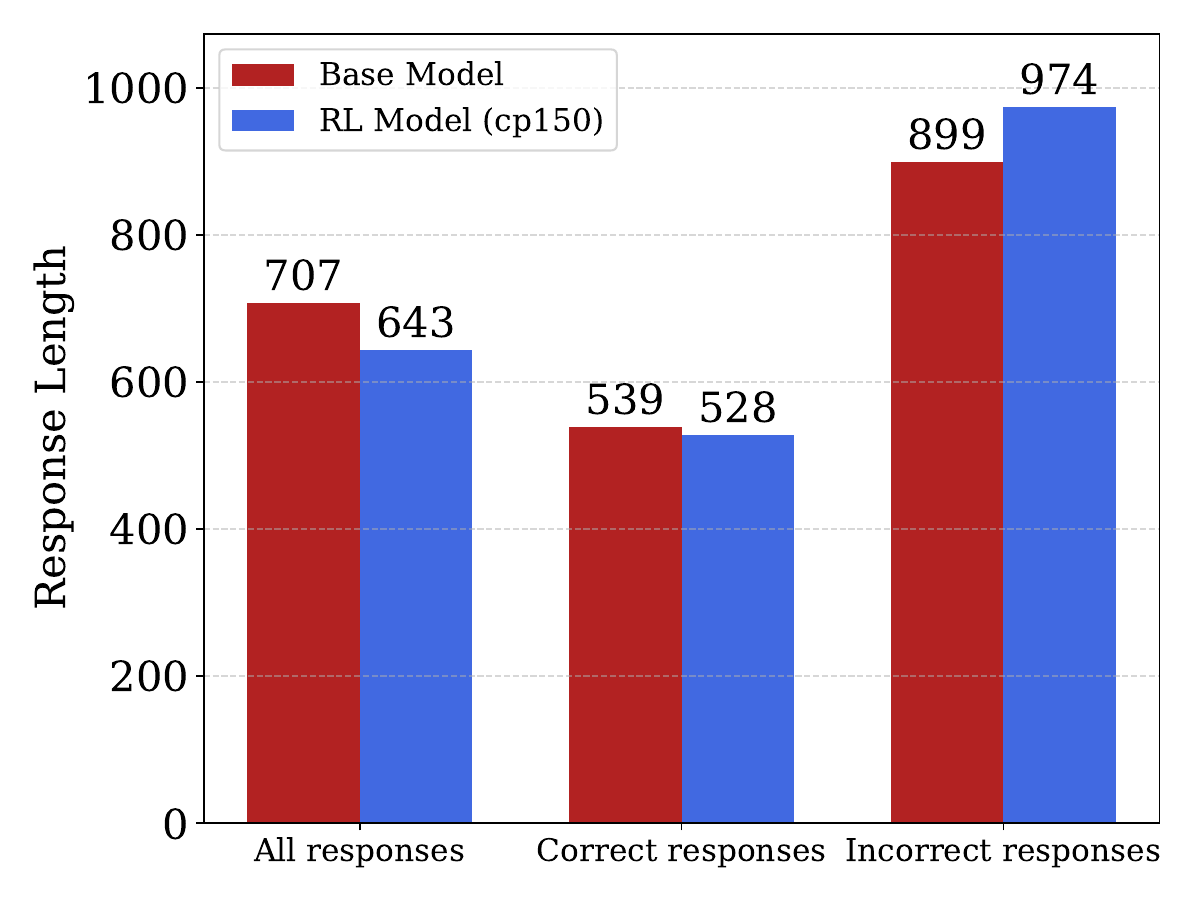}
    \caption{Mean response length grouped by correctness.}
    \label{fig:response-length-correctness}
  \end{subfigure}
  \hfill
  \begin{subfigure}[t]{0.6\textwidth}
    \centering
    \includegraphics[width=\linewidth]{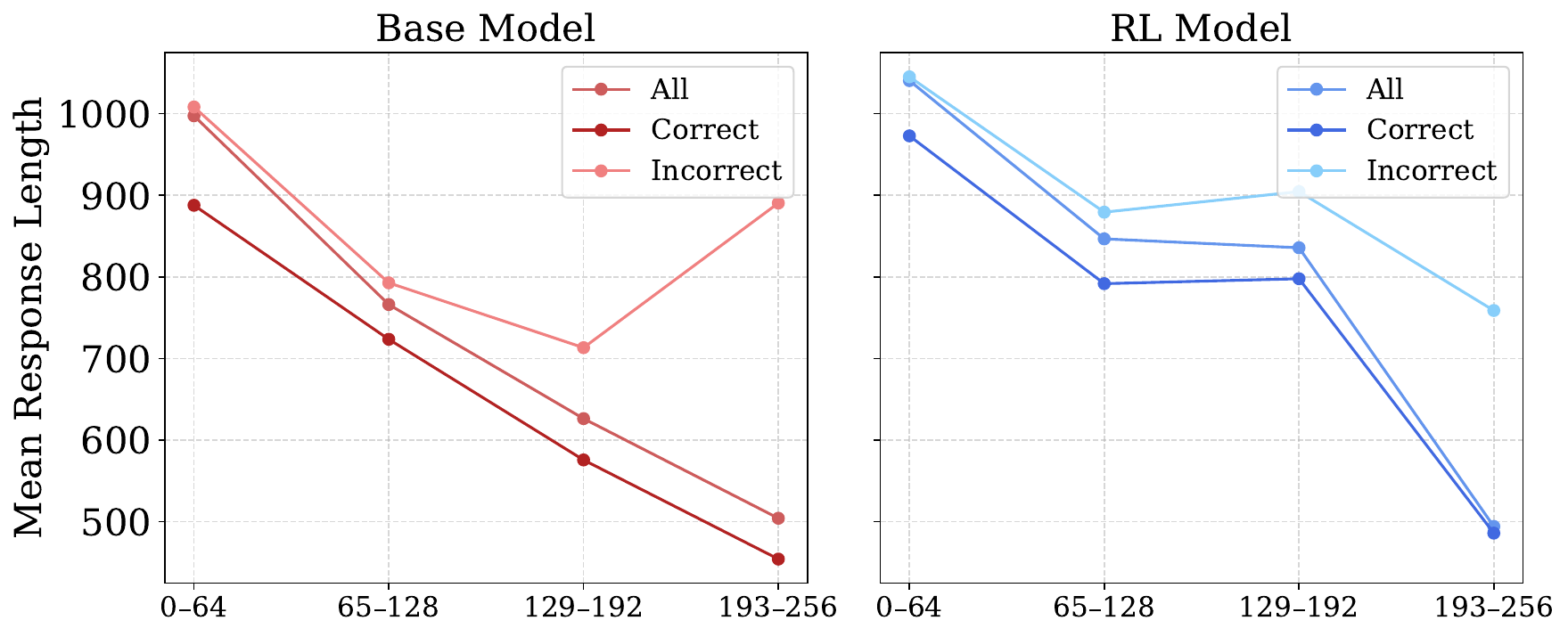}
    \caption{Mean response length stratified by difficulty and correctness.}
    \label{fig:response-length-difficulty}
  \end{subfigure}
  \caption{Comparison of response lengths between base and RL models for Qwen-2.5-1.5B-Math. RLVR did not increase verbosity, and correct answers tended to be shorter.}
  \label{fig:appendix_response_length}
\end{figure}

\begin{figure}[H]
  \centering
  \begin{subfigure}[t]{0.3\textwidth}
    \centering
    \includegraphics[width=\linewidth]{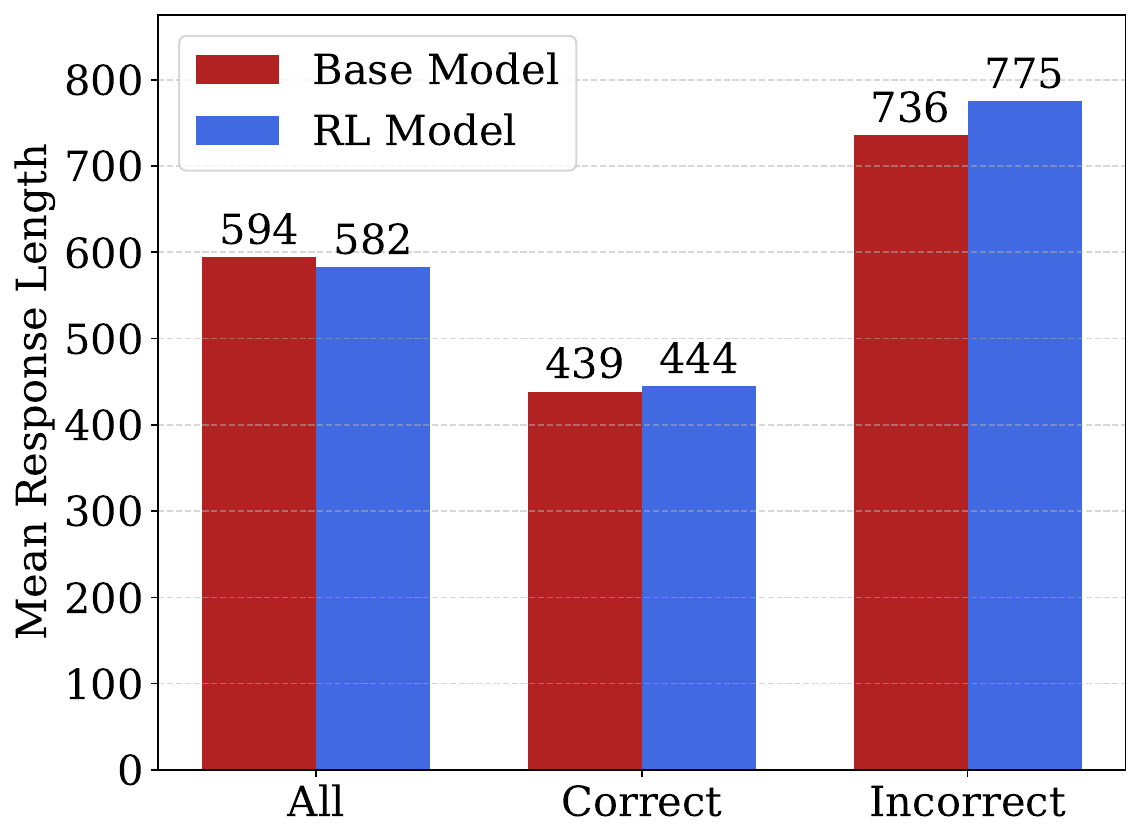}
    \caption{Mean response length grouped by correctness.}
    \label{fig:response-length-correctness_2}
  \end{subfigure}
  \hfill
  \begin{subfigure}[t]{0.6\textwidth}
    \centering
    \includegraphics[width=\linewidth]{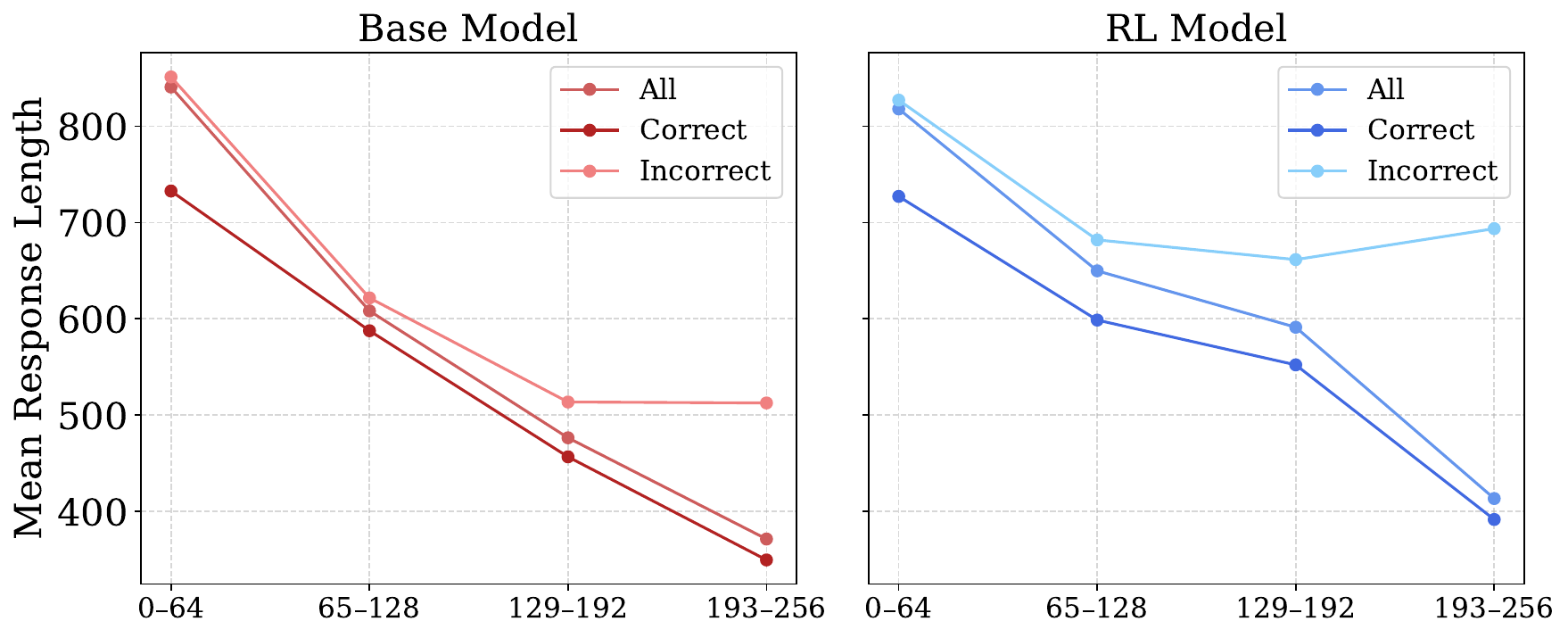}
    \caption{Mean response length stratified by difficulty and correctness.}
    \label{fig:response-length-difficulty_2}
  \end{subfigure}
  \caption{Comparison of response lengths between base and RL models for Qwen-2.5-3B. RLVR did not increase verbosity, and correct answers tended to be shorter.}
  \label{fig:appendix_response_length_2}
\end{figure}

\begin{figure}[H]
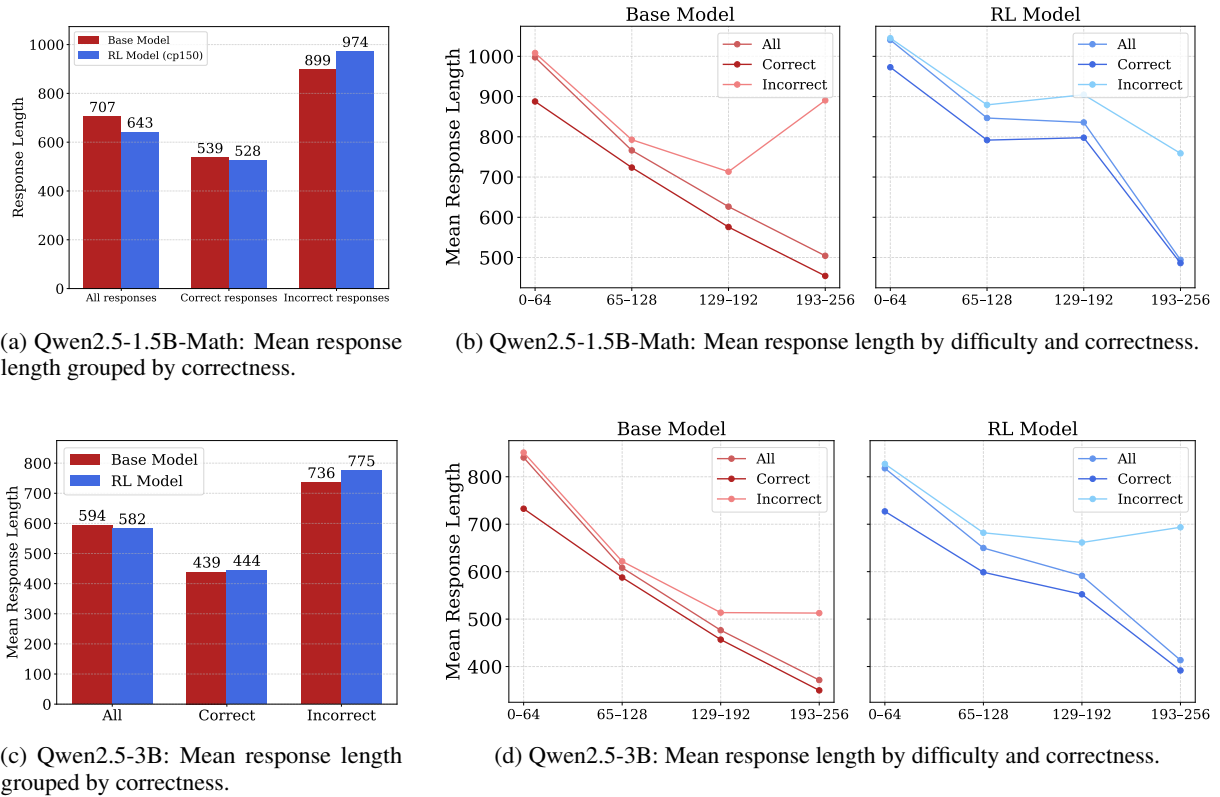

  \centering
  \begin{subfigure}[t]{0.3\textwidth}
    \centering
    \includegraphics[width=\linewidth]{images/response_length_count_1.5b.pdf}
    \caption{Qwen2.5-1.5B-Math: Mean response length grouped by correctness.}
    \label{fig:response-length-correctness-1.5B}
  \end{subfigure}
  \hfill
  \begin{subfigure}[t]{0.6\textwidth}
    \centering
    \includegraphics[width=\linewidth]{images/response-length-difficulty_1.5b.pdf}
    \caption{Qwen2.5-1.5B-Math: Mean response length by difficulty and correctness.}
    \label{fig:response-length-difficulty-1.5B}
  \end{subfigure}

  \vspace{1em}

  \begin{subfigure}[t]{0.3\textwidth}
    \centering
    \includegraphics[width=\linewidth]{images/response-length-plot.pdf}
    \caption{Qwen2.5-3B: Mean response length grouped by correctness.}
    \label{fig:response-length-correctness-3B}
  \end{subfigure}
  \hfill
  \begin{subfigure}[t]{0.6\textwidth}
    \centering
    \includegraphics[width=\linewidth]{images/response-length-difficulty.pdf}
    \caption{Qwen2.5-3B: Mean response length by difficulty and correctness.}
    \label{fig:response-length-difficulty-3B}
  \end{subfigure}

  \caption{Comparison of response lengths between base and RLVR-trained models across model sizes and difficulty levels. Top row: Qwen2.5-1.5B-Math; bottom row: Qwen2.5-3B. Left: Mean response length grouped by correctness. Right: Mean response length further stratified by difficulty. In both models, RLVR did not increase response length, and correct responses tended to be more concise.}
  \label{fig:appendix_response_length_combined}
\end{figure}

For both 1.5B and 3B models, we generated 256 responses per MATH500 question from both the base and RL models and computed mean response lengths. We also separated responses by correctness. As shown in Figure~\ref{fig:appendix_response_length_combined}, there was no substantial difference in average length between the two models. In both cases, correct responses were consistently shorter than incorrect ones.

To control for the correlation between question difficulty and correctness, we grouped questions into four bins based on how many of the 256 base model responses were correct—higher bins indicating easier questions. Within each bin, we compared mean response lengths by correctness. As shown in the figure, both models exhibited the same trend: correct responses were consistently shorter, and overall response lengths remained similar, indicating that RLVR did not increase response length.

\subsubsection{Reflection-Related Keywords} 
\begin{figure}[H]
  \centering
  \begin{subfigure}[t]{0.33\textwidth}
    \centering
    \includegraphics[width=\textwidth]{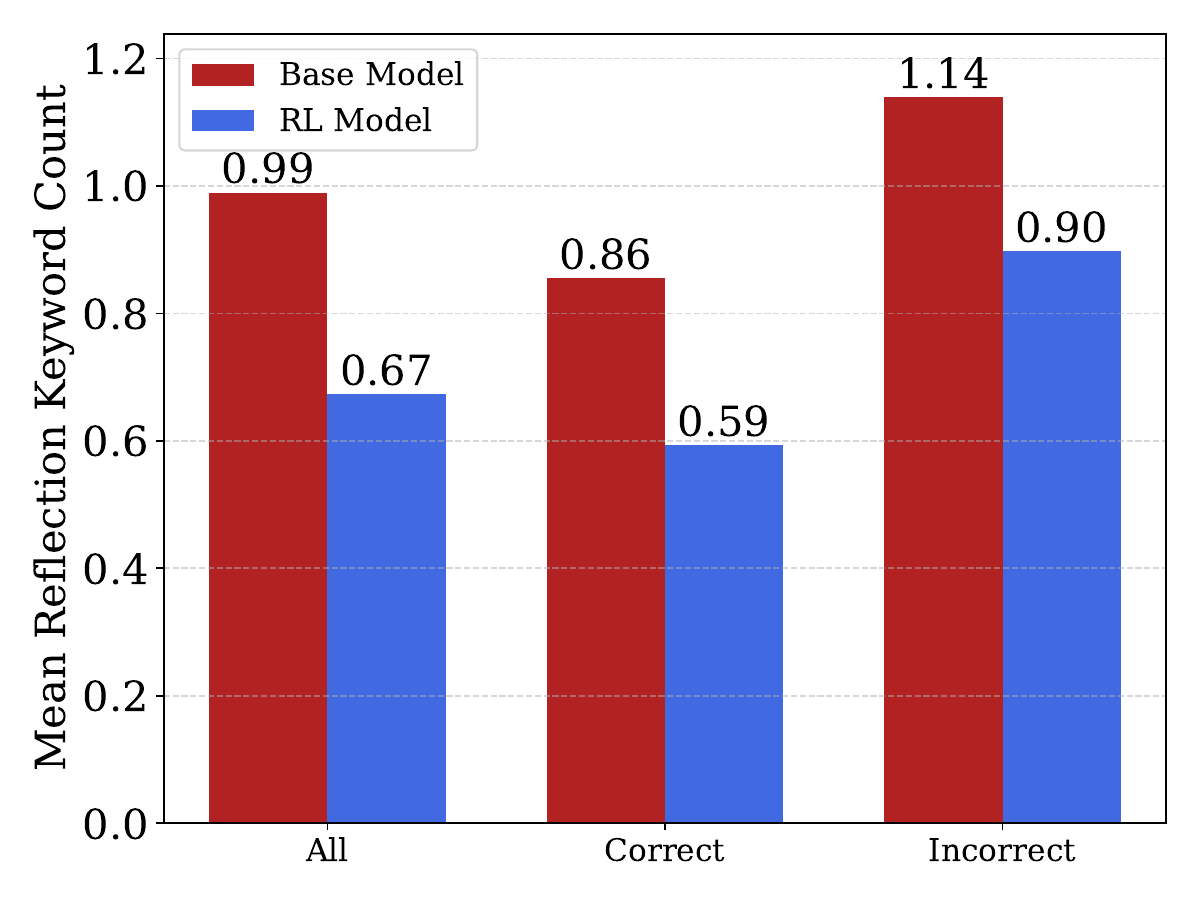}
    \caption{Qwen2.5-1.5B-Math: Mean count of reflection keywords grouped by correctness.}
    \label{fig:reflection-keyword-correctness-1.5B}
  \end{subfigure}
  \hfill
  \begin{subfigure}[t]{0.64\textwidth}
    \centering
    \includegraphics[width=\textwidth]{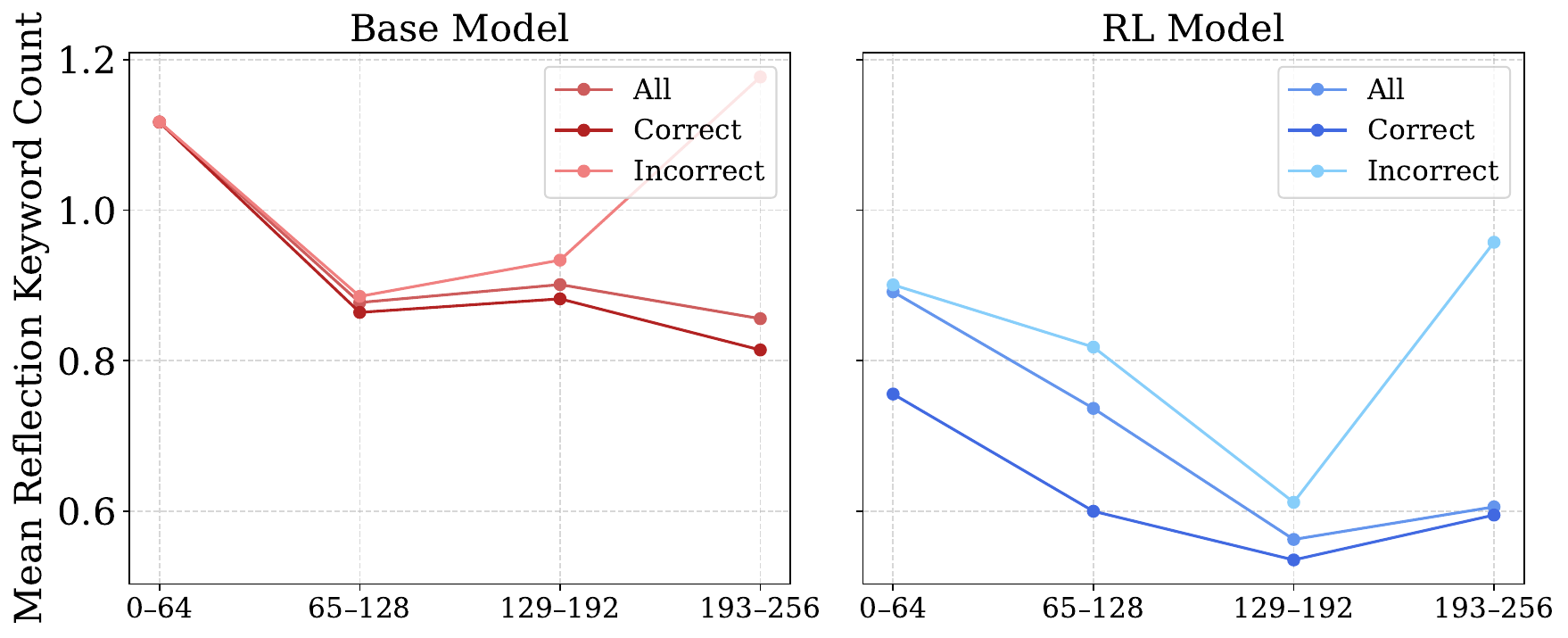}
    \caption{Qwen2.5-1.5B-Math: Reflection keyword frequency by difficulty and correctness.}
    \label{fig:reflection-keyword-difficulty-1.5B}
  \end{subfigure}

  \vspace{1em}

  \begin{subfigure}[t]{0.33\textwidth}
    \centering
    \includegraphics[width=\textwidth]{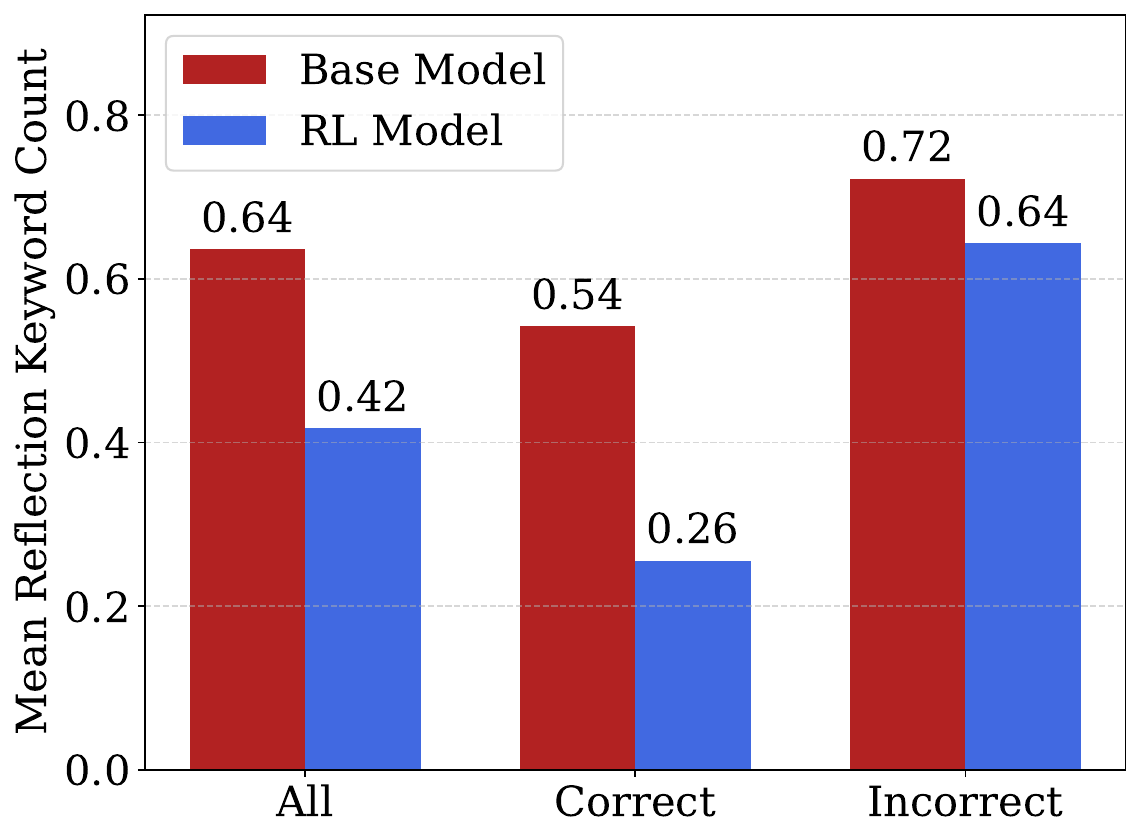}
    \caption{Qwen2.5-3B: Mean count of reflection keywords grouped by correctness.}
    \label{fig:reflection-keyword-correctness-3B}
  \end{subfigure}
  \hfill
  \begin{subfigure}[t]{0.64\textwidth}
    \centering
    \includegraphics[width=\textwidth]{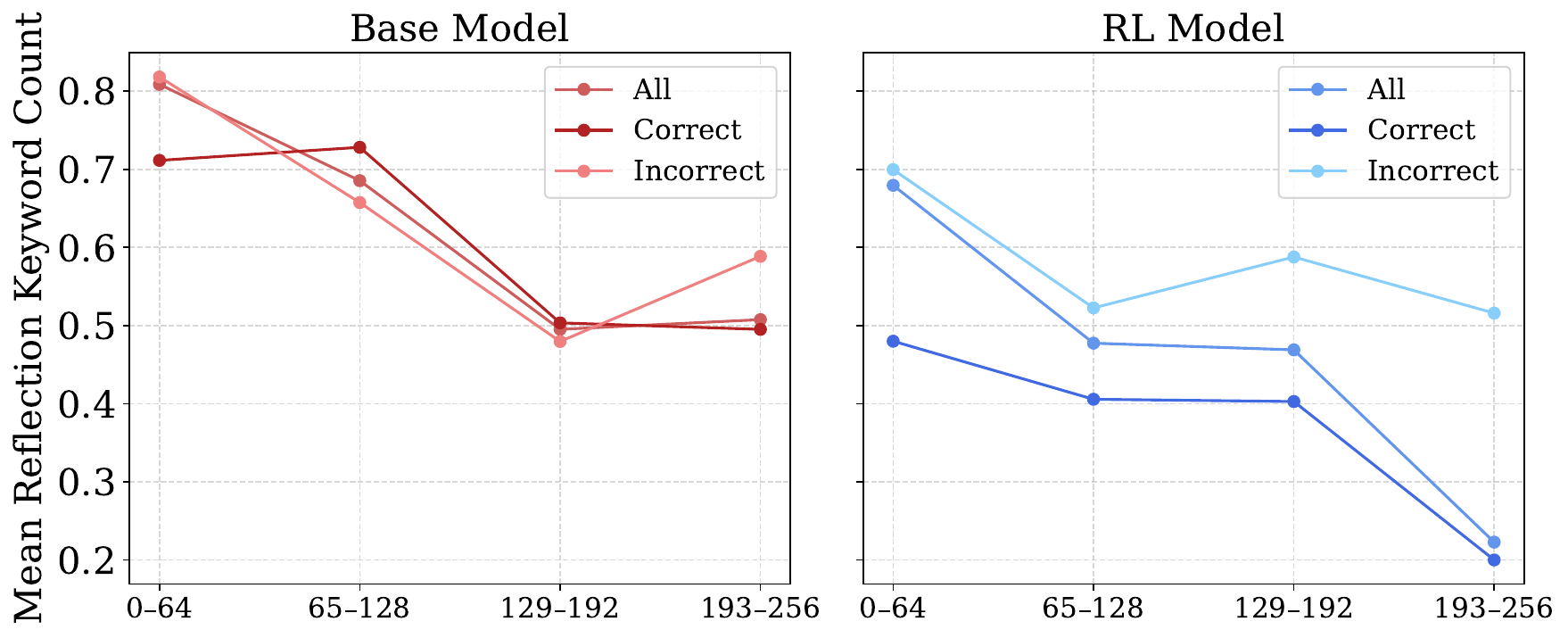}
    \caption{Qwen2.5-3B: Reflection keyword frequency by difficulty and correctness.}
    \label{fig:reflection-keyword-difficulty-3B}
  \end{subfigure}

  \caption{Reflection-related keyword analysis across base and RLVR-trained models. Top row: Qwen2.5-1.5B-Math; bottom row: Qwen2.5-3B. Left: Mean count of reflection-related keywords, grouped by correctness. Right: Keyword frequency stratified by question difficulty and correctness. Across both models, RLVR-trained responses consistently contain fewer reflective phrases.}
  \label{fig:appendix_keyword_count_combined}
\end{figure}

Prior work suggests that RLVR elicits more non-linear reasoning in model outputs~\cite{deepseek2025r1, gandhi2025cognitive}. To test this, we analyzed the presence of predefined reflection-related phrases. The full list of the phrases are available in Table \ref{tab:reflection_keywords}.

\begin{table}[htbp]
  \centering
  \begin{tabular}{lll}
    \toprule
    \textbf{Reflection-Related Keywords} & & \\
    \midrule
    actually              & aha                 & alternatively \\
    another approach      & checking our work   & correction: \\
    different method      & double-check        & hmm \\
    however               & I made a mistake    & I need to reconsider \\
    I realize             & let me recalculate  & let me think \\
    let’s check           & let’s reconsider    & looking back \\
    make sure             & ok                  & on second thought \\
    retracing             & to be sure          & to confirm \\
    verify                & wait                & we could also \\
    \bottomrule
  \end{tabular}
  \vspace{0.5em}
  \caption{List of reflection-related phrases used for qualitative analysis.}
  \label{tab:reflection_keywords}
\end{table}

Using the same setup as the response length analysis, we examined 256 responses per question across all MATH500 test questions for both base and RL models. For each response, we counted occurrences of reflection keywords and stratified results by correctness and difficulty.

As shown in Figure~\ref{fig:appendix_keyword_count_combined}, the RL model exhibited substantially fewer reflection keywords than the base model. The figure further shows that, while the base model showed little variation across correctness levels, the RL model consistently used fewer reflection phrases in correct answers across all difficulty bins. These responses were generally more direct and less exploratory.


\subsection{QwQ-32B Capability Experiment} \label{appendix:qwq}

\begin{figure}[htbp]
  \centering
  \includegraphics[width=0.50\textwidth]{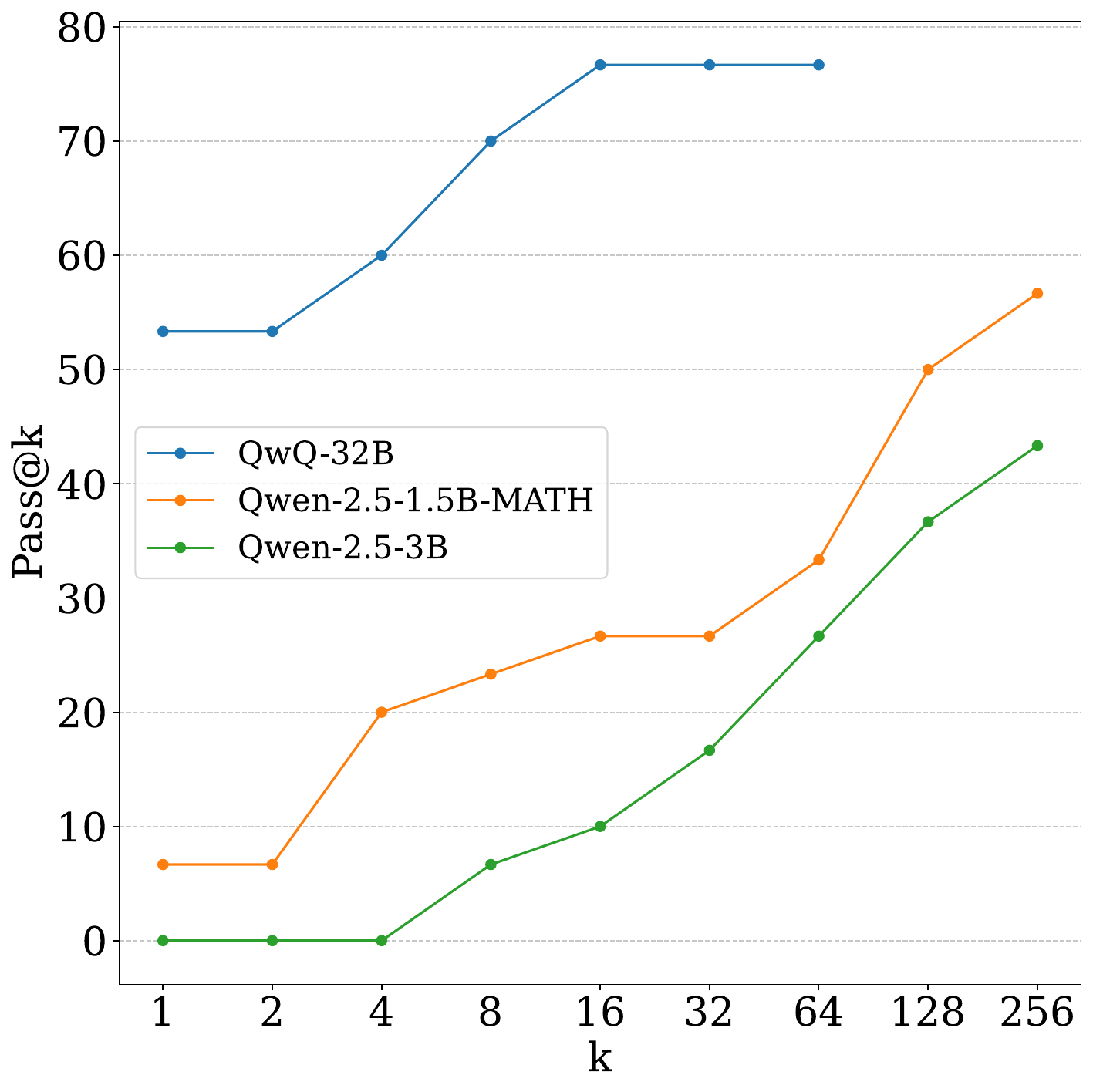}
  \caption{Pass@$k$ results of QwQ-32B, Qwen2.5-3B-Math, and Qwen2.5-1.5B-Math on AIME 25}
  \label{fig:qwq_pass@k}
\end{figure}

In Section~\ref{sec:section6}, we selected QwQ-32B as the teacher model for our reasoning-only distillation experiment. To ensure a fair test of whether distillation can improve capability without introducing new knowledge, the teacher must have higher capability than the student models—Qwen2.5-3B and Qwen2.5-1.5B-Math.

To validate this, we conducted a pass@$k$ evaluation on AIME 25 using 64 responses per question from QwQ-32B, and compared the results with the two student models. As shown in Figure~\ref{fig:qwq_pass@k}, QwQ-32B consistently outperforms both students across all $k$ values, with no sign of convergence. Notably, its pass@64 score reached 76.7\%, compared to just 43.3\% and 56.7\% at pass@256 for Qwen2.5-3B and Qwen2.5-1.5B-Math, respectively. These results confirm that QwQ-32B has substantially higher capability, making it a suitable teacher model for our distillation setup.



\newpage 
\subsection{Teacher Distillation Pass@$k$ results} \label{appendix:distillation pass@k}

As discussed in Section \ref{sec:section6}, we conducted the pass@$k$ on AIME 25 and MATH 500 for both Qwen2.5-1.5B-Math and Qwen2.5-3B and each their 2 distilled variants. The results are shown below in Figure \ref{fig:distillation_pass_at_k_stacked}.

    
    

\begin{figure}[htbp]
  \centering
  \includegraphics[width=0.9\textwidth]{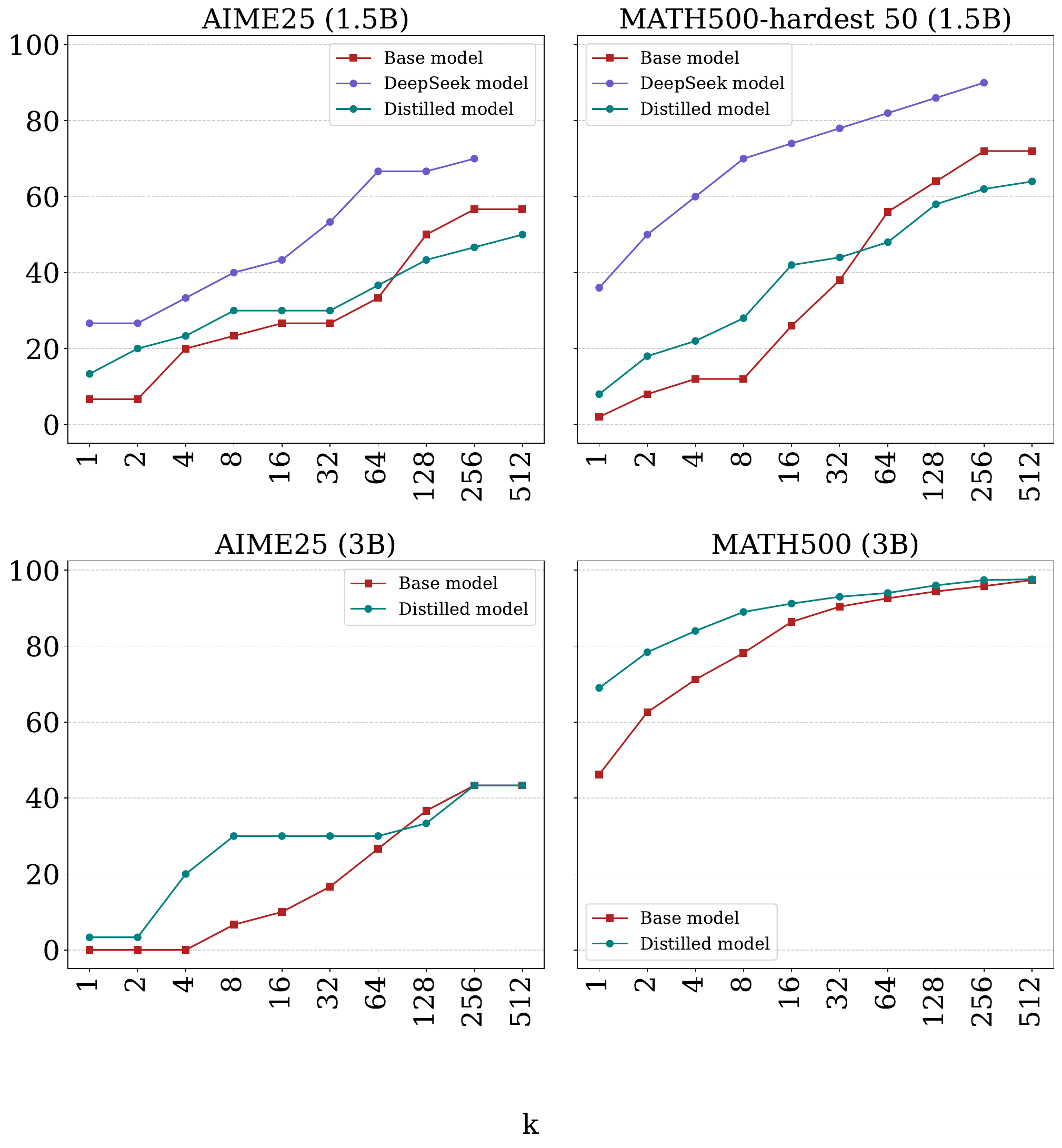}
  \caption{Pass@$k$ comparisons across AIME25 and MATH500 datasets for both 1.5B (top) and 3B (bottom) models and their distillation-trained variants. For the MATH500 results of the 1.5B models, we show performance on the 50 questions with the lowest base-model success rates to better highlight the differences.}
  \label{fig:distillation_pass_at_k_stacked}
\end{figure}

\newpage

\subsection{Qwen2.5-Math-1.5B Training Details}\label{appendix:1.5B-math}

In this paper, we used two models as base models: Qwen2.5-1.5B-Math and Qwen2.5-3B. For the RLVR-trained version of the 1.5B model, we used Qwen2.5-Math-1.5B-Oat-Zero\footnote{\url{https://huggingface.co/sail/Qwen2.5-Math-1.5B-Oat-Zero}}, a publicly available model trained by \citeauthor{liu2025understanding}. According to their report, the model was trained with Dr.GRPO \cite{liu2025understanding}, a variant of the GRPO algorithm \cite{shao2024deepseekmath} designed to remove response length and question difficulty biases. The model was trained on questions from level 3 to 5 from the MATH training set. For the 3B model, we performed RLVR training ourselves. Training details are shown right below in Appendix \ref{appendix:3b}

\subsection{Qwen2.5-3B RLVR Training Details} \label{appendix:3b}

For RLVR training of Qwen2.5-3B, we used the \texttt{GRPOTrainer} from the \texttt{TRL\footnote{\url{https://github.com/huggingface/trl}}} library, which implements the standard GRPO algorithm. The model was trained on the full MATH training set, consisting of 7,500 questions.

\subsubsection{Prompt Setting}

Prior work has shown that the performance of smaller models can be sensitive to prompt design \cite{hochlehnert2025sober, liu2025understanding}. Following \citeauthor{liu2025understanding}, we evaluated three prompt formats, as listed below. We ultimately adopted Template 3 (question only), which yielded the best performance.

\begin{tcolorbox}[title=Prompt Templates, colback=blue!10, colframe=blue!50, fonttitle=\bfseries]

\textbf{Template 1 (R1 template)}  
A conversation between User and Assistant. The User asks a question, and the Assistant solves it. The Assistant first thinks about the reasoning process in the mind and then provides the User with the answer. The reasoning process is enclosed within \texttt{<think> </think>} and the answer is enclosed within \texttt{<answer> </answer>} tags.  
\texttt{User: \{question\}}  
\texttt{Assistant: <think> reasoning here </think> <answer> answer here </answer>}

\vspace{1em}
\textbf{Template 2 (Qwen-Math template)}  
\texttt{<|im start|>system}  
Please reason step by step, and put your final answer within \textbackslash boxed\{\}.  
\texttt{<|im end|>}  
\texttt{<|im start|>user}  
\{question\}  
\texttt{<|im end|>}  
\texttt{<|im start|>assistant}

\vspace{1em}
\textbf{Template 3 (Question only)}  
\{question\}
\end{tcolorbox}

\subsubsection{Reward Function}

We adopted a minimalistic reward setting. A response received a reward of 1 if it contained the correct final answer, and -1 otherwise. Answer verification was performed using the \texttt{math\_verify}\footnote{\url{https://github.com/huggingface/Math-Verify}} package.

\[
R(q, a, r) =
\begin{cases}
1 & \text{if the response $r$ to question $q$ matches the ground truth answer $a$} \\
-1 & \text{otherwise}
\end{cases}
\]

\newpage
\subsubsection{RLVR Training Hyperparameters}

Table~\ref{tab:grpo-hyperparams} summarizes the key hyperparameters used in RLVR training for the Qwen2.5-3B model.

\begin{table}[H]
\centering
\begin{tabularx}{\linewidth}{@{}lX@{}}
\toprule
\textbf{Hyperparameter} & \textbf{Value} \\
\midrule
Optimizer & AdamW \\
Learning rate scheduler & Constant \\
Maximum token length & 4000 \\
Temperature & 0.9 \\
Top-$p$ & 1.0 \\
Top-$k$ & 50 \\
Number of generations (per question) & 10 \\
Global batch size & 4 (per device) $\times$ 7 (GPUs) $\times$ 10 (accumulation) = 280 \\
Learning rate & $1 \times 10^{-6}$ \\
Gradient clipping (max grad norm) & 0.1 \\
Number of gradient steps & 225 \\
Warmup steps & 20 \\
Mixed precision & bf16 \\
\bottomrule
\end{tabularx}
\caption{Key hyperparameters used for RLVR training of Qwen2.5-3B.}
\label{tab:grpo-hyperparams}

\end{table}

\subsubsection{Training Progress and Evaluation} \label{reward_graph}

\begin{figure}[H]
  \centering
  \includegraphics[width=0.6\columnwidth]{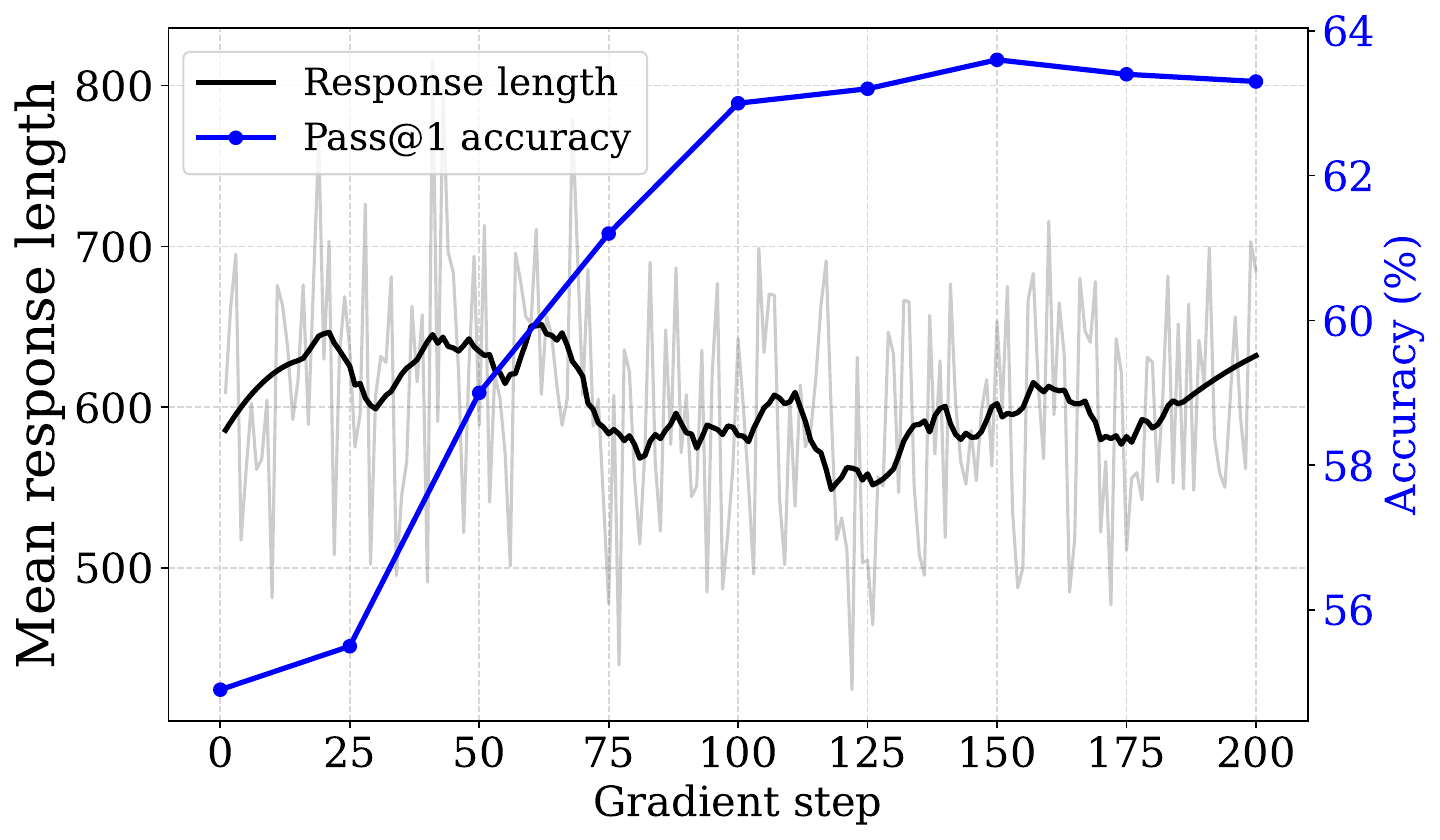}
  \caption{Change in response length (token counts) over training and accuracy on MATH 500 across checkpoints.}
  \label{fig:training_plot}
\end{figure}

During RLVR training, we evaluated the model every 25 gradient steps. To ensure statistical robustness, we followed the recommendation of \citeauthor{hochlehnert2025sober}~\cite{hochlehnert2025sober} and sampled responses 10 times per checkpoint, reporting the mean accuracy. For each evaluation, we used a temperature of 0.8 and a top-$p$ of 0.9.

As shown in Figure~\ref{fig:training_plot}, accuracy peaked at step 150, reaching 63.6\%, and then plateaued. We selected this checkpoint as the RL model used throughout our experiments. The figure also shows the average response length over training. As discussed in Section~\ref{qualitative_analysis}, response length remained stable, showing no significant growth.

\newpage
\subsection{Distillation Training Hyperparameters} \label{appendix:sft_hyperparams}

For all the self-distillation experiments in Section~\ref{sec:self-distillation} and teacher distillation in Section~\ref{sec:section6}, we used the supervised fine-tuning (SFT) hyperparameters listed in Table~\ref{tab:sft-hyperparams}.

\begin{table}[H]
\centering
\begin{tabularx}{\linewidth}{@{}lX@{}}
\toprule
\textbf{Hyperparameter} & \textbf{Value} \\
\midrule
Optimizer & AdamW \\
Learning rate scheduler & Constant \\
Weight decay & $1 \times 10^{-4}$ \\
Warmup steps & 25 \\
Max sequence length & 32,768 \\
Global batch size &  4 \\
Mixed precision & bf16 \\
\bottomrule
\end{tabularx}
\caption{Key hyperparameters used for supervised fine-tuning in distillation experiments.}
\label{tab:sft-hyperparams}
\end{table}


\subsection{Response Generation Details}

We used vLLM\footnote{\url{https://docs.vllm.ai}} library \cite{kwon2023efficient} for response generation  and \texttt{math\_verify}\footnote{\url{https://github.com/huggingface/Math-Verify}} package for response grading.

\noindent
We used temperature 0.9, top-$p$ of 1.0, and top-$k$ of 50 for all models, except where noted below. These settings were chosen to ensure response diversity. Unless otherwise specified, we used the question-only template (Template 3).

\noindent
For Qwen2.5-Math-1.5B-Oat-Zero, we used the same sampling hyperparameters but followed the Qwen prompt format (Template 2), as recommended in the user guideline.\footnote{\url{https://huggingface.co/sail/Qwen2.5-Math-1.5B-Oat-Zero}}

\noindent
For QwQ-32B, we used temperature 0.6, top-$p$ 0.95, and top-$k$ 50. We followed the R1 prompt template (Template 1), as recommended in the user guideline.\footnote{\url{https://huggingface.co/Qwen/QwQ-32B\#usage-guidelines}}

\noindent
For DeepSeek-R1-Distill-Qwen-1.5B, we used temperature 0.6, top-$p$ 0.95, and top-$k$ 50. We followed the R1 prompt template (Template 1), as recommended in the user guideline.\footnote{\url{https://huggingface.co/deepseek-ai/DeepSeek-R1-Distill-Qwen-1.5B\#usage-recommendations}}





\twocolumn

\end{document}